\documentclass[runningheads]{llncs}

\usepackage{eccv}

\usepackage{eccvabbrv}

\usepackage{graphicx}
\usepackage{booktabs}

\usepackage[accsupp]{axessibility}  % Improves PDF readability for those with disabilities.

\usepackage{hyperref}
\usepackage{listings}
\usepackage{stfloats}
\usepackage{orcidlink}
\usepackage[dvipsnames]{xcolor}
\usepackage{multirow}
\usepackage{tabularx}
\usepackage{graphicx}
\usepackage{makecell}
\usepackage{siunitx}
\usepackage{booktabs}
\usepackage{hhline} 

\usepackage[dvipsnames]{xcolor}
\usepackage{colortbl}
\usepackage{amsmath}
\usepackage{amssymb}
\usepackage{booktabs}
\usepackage{caption} % Add this if you want to customize captions
\usepackage{arydshln}

\usepackage{bbding}
\usepackage{pifont}

\usepackage{amssymb} % for \checkmark
\usepackage{xcolor} 
\definecolor{hh}{RGB}{255,0,0}

\definecolor{ml}{RGB}{165, 42, 42}

\definecolor{wj}{RGB}{0,128,200}
\definecolor{wj2}{RGB}{0,120,200}

\definecolor{hr}{RGB}{0,128,0}

\definecolor{mygray}{gray}{.9}

\usepackage[table]{xcolor}
\definecolor{Best}{HTML}{D6F5D6}   % 淡绿色：最佳
\definecolor{Second}{HTML}{E8F0FE} % 淡蓝色：次佳

\newcommand{\cmark}{\ding{51}} % 勾号
\newcommand{\xmark}{\ding{55}} % 叉号
\usepackage{pifont}

\newcommand{\bestdown}[1]{\cellcolor{Best}\textbf{#1}}
\newcommand{\seconddown}[1]{\cellcolor{Second}#1}
\newcommand{\bestup}[1]{\cellcolor{Best}\textbf{#1}}
\newcommand{\secondup}[1]{\cellcolor{Second}#1}

\usepackage{wrapfig}

\begin{document}

% ---------------------------------------------------------------
% TODO REVIEW: Replace with your title
\title{GRE-Diff: Gaussian Room Embeddings for Structured Layout Diffusion} 

% TODO REVIEW: If the paper title is too long for the running head, you can set
% an abbreviated paper title here. If not, comment out.
\titlerunning{GRE-Diff}

% TODO FINAL: Replace with your author list. 
% Include the authors' OCRID for the camera-ready version, if at all possible.
\author{Jing Wang\inst{1}\orcidlink{0000-0002-9792-7165} \and
Haoran Xiong\inst{1}\orcidlink{0009-0005-7961-7648} \and
Zihao Yan\inst{1}\orcidlink{0000-0003-2160-9109} \and
Minglun Gong\inst{2}\orcidlink{0000-0001-5820-5381} \and \\
Hui Huang\inst{1}\orcidlink{0000-0003-3212-0544}\thanks{Corresponding author.}}

% TODO FINAL: Replace with an abbreviated list of authors.
\authorrunning{J. Wang, H. Xiong, Z. Yan, M. Gong, and H. Huang}
% First names are abbreviated in the running head.
% If there are more than two authors, 'et al.' is used.

% TODO FINAL: Replace with your institution list.
\institute{
Guangdong Provincial Key Laboratory of Visual Media and Multidimensional Intelligence, CSSE, Shenzhen University, China\\
%\email{hhzhiyan@gmail.com}
%\and
%VIVO Mobile Communication Co., Ltd., China \\
%\email{mr.salingo@gmail.com}
\and
School of Computer Science, University of Guelph, Canada \\
%\email{gongml@gmail.com}
}

\maketitle

\begin{abstract}
  Designing functional and aesthetically coherent floor plans requires exploring a vast space of possible room arrangements, a task that quickly becomes overwhelming for human designers. In this paper, we propose \emph{GRE-Diff}, a controllable and interactive diffusion-based framework that automates the creation and editing of apartment floor plans under user-specified constraints. 
  By combining AI-generated suggestions with real-time, human-in-the-loop editing, the system enables %users to define boundary shapes, room types, and layout requirements through either natural-language instructions or GUI-based interaction. 
  users to specify room types, room counts, boundary shapes, and editing operations through LLM-parsed instructions or GUI-based interaction.
  It then generates a diverse set of plausible and well-structured designs for refinement. 
  At the core of our approach is Gaussian Room Embedding (GRE), a continuous latent representation that models each room as a spatial Gaussian distribution capturing its location and extent.  
  Extensive experiments on the RPLAN dataset show that \emph{GRE-Diff} produces high-quality, constraint-aware, and editable polygonal layouts, offering a practical step toward bridging AI-driven automation and human creativity in spatial design.
  
  \keywords{Floor Plan \and Layout Generation \and Controllable Editing}
\end{abstract}

\section{Introduction}\label{sec:intro}

Designing functional and well-structured floor plans is inherently challenging, as architects must iteratively refine layouts to satisfy complex spatial and regulatory constraints~\cite{sully2015interior,weber2022automated}. 
The resulting design space is highly combinatorial, involving numerous topological and geometric configurations that are difficult to explore exhaustively.
Recent advances in artificial intelligence have sought to address this challenge through data-driven generative models, ranging from graph-based reasoning~\cite{hu2020Graph2Plan,nauata2020housegan,para2021generative,upadhyay2022flnet,nauata2103housegan++,he2022iplan,wang2023automated,zhang2024MaskPLAN,dupty2024constrained,tang2024graphTransGAN} to diffusion-based layout synthesis ~\cite{shabani2023housediffusion,hong2024cons2plan,qin2024chathousediffusion,ploennigs2024automating,hu2025gsdiff,xu2025FloorplanDiffusion}. 
These methods demonstrate the promise of AI to explore complex spatial design spaces beyond human capabilities.
% Graph: hu2020graph2plan,nauata2020housegan,para2021generative,upadhyay2022flnet,nauata2103housegan++,he2022iplan,wang2023automated,zhang2024MaskPLAN,dupty2024constrained,tang2024graphTransGAN
% Diffusion: shabani2023housediffusion,hong2024cons2plan,qin2024chathousediffusion,ploennigs2024automating,hu2025gsdiff,xu2025FloorplanDiffusion

However, enabling controllable generation and iterative refinement within a unified generative framework remains challenging.
Unlike pixel-level image synthesis, floor plans are structured geometric systems where rooms are mutually constrained by boundary validity,
spatial adjacency, and functional relationships. 
This structure introduces three key challenges.
First, controllable generation under spatial constraints requires simultaneously maintaining geometric validity, functional consistency, and
inter-room relationships.
Second, robust editing is challenging because modifying one room often propagates structural changes to adjacent spaces, making it difficult to preserve global layout coherence.
Third, existing representations rely on discrete masks or vertex sequences, which are sensitive to ordering and coordinate perturbations,
leading to unstable optimization and geometrically brittle layouts.
These observations suggest that the fundamental difficulty lies in how room geometry is represented within generative frameworks.
To overcome the limitations of discrete room representations, %we seek a representation that is continuous, permutation-invariant, and geometrically robust.
we seek a representation that is continuous, less sensitive to valid vertex-order variations, and geometrically robust.

Inspired by probabilistic embedding models~\cite{vilnis2014word} and spatially controllable generative representations~\cite{epstein2022blobgan, li2025blobctrl}, we introduce a probabilistic latent representation for room geometry.
Specifically, we associate each room with a Gaussian spatial embedding that captures its approximate location and spatial extent.
We refer to this representation as the \textbf{Gaussian Room Embedding (GRE)}.
Building upon this representation, we propose \emph{GRE-Diff}, a Gaussian-guided diffusion framework for controllable floor plan generation and editing. 
Given user-specified constraints, the framework predicts Gaussian room embeddings that guide a diffusion process to generate coherent polygonal layouts. 
Operating in the continuous GRE space enables unified layout generation and iterative editing within a single diffusion model. 
Furthermore, the system supports both natural-language instructions and interactive graphical editing, allowing users to refine layouts while maintaining global spatial coherence.

\par
Our main contributions are summarized as follows:
\begin{itemize}
    %\item We introduce Gaussian Room Embeddings, a continuous and permutation-invariant probabilistic representation that reformulates room geometry for structured layout generation.
    \item We introduce Gaussian Room Embeddings, a continuous probabilistic room representation that summarizes room location and spatial extent, reducing sensitivity to valid vertex-order variations in polygonal layouts.
    \item We propose \emph{GRE-Diff}, a Gaussian-guided diffusion framework that unifies floor plan generation and editing by operating in the GRE latent space and decoding coherent polygonal layouts.
    \item We design a dual-path conditioning architecture that captures both semantic cues and geometric constraints, enabling controllable and structure-preserving layout synthesis and refinement.
\end{itemize}

\section{Related Work}
\label{sec:rw}

We mainly review the learning-based methods as they are closely related to our research. 
Based on the type of input conditions, we categorize existing studies into three groups: boundary-constrained, bubble-constrained, and text-constrained floor plan generation.

\subsection{Boundary-constrained generation}
Boundary constraints are commonly used in floor plan generation to ensure that the layout adheres to predefined building boundaries or exterior walls, which can be specified by designers or users. 
Early methods, such as variational autoencoders (VAEs)~\cite{kingma2013auto,bao2017cvae}, approached floor plan generation as a raster image synthesis problem, where boundary constraints guided the placement of rooms.
RPLAN~\cite{wu2019Data-Driven} enhanced layout accuracy and efficiency by generating floor plans from raster images.
iPLAN~\cite{he2022iplan} mimicked the human design process by iteratively predicting room locations and partitions.
WallPlan~\cite{sun2022WallPlans} employed two alternating modules to predict wall and room functions, but relied on post-processing to obtain complete layouts, which limited output diversity.
MaskPLAN~\cite{zhang2024MaskPLAN} improved spatial coherence by refining individual elements and decoupling geometric interdependencies. 
However, boundary-based methods primarily focus on geometric validity, often overlooking the semantic relationships among rooms. This limitation led to the emergence of bubble-constrained methods, which introduce higher-level spatial reasoning.

\subsection{Bubble-constrained generation}
Another line of research~\cite{liu2022end, upadhyay2022flnet} leverage bubble diagrams as abstract spatial constraints, providing room connectivity, adjacency, and relative proportions before geometric realization, usually combined with graph neural networks (GNNs)~\cite{hu2020Graph2Plan,dupty2024constrained,wang2023automated} and generative adversarial networks (GANs)~\cite{nauata2020housegan,  nauata2103housegan++,tang2024graphTransGAN}.
More recently, vision transformers~\cite{vaswani2017attention,para2021generative} and diffusion models~\cite{ploennigs2024automating,dhariwal2021diffusion} have been explored for layout synthesis.
%Hu \etal~\cite{hu2020Graph2Plan} proposed Graph2Plan, a framework that takes sparse user constraints and bubble diagrams as input to generate coherent floor plans, using graph structures to guide layout synthesis. 
Graph2Plan~\cite{hu2020Graph2Plan} used sparse user constraints and bubble diagrams with GNNs to generate coherent floor plans.
HouseGAN++~\cite{nauata2103housegan++} iteratively generated floor plans using a graph-constrained GAN, effectively transforming input graphs into coherent and plausible layouts.
%ActFloor-GAN~\cite{wang2021actfloor} was one of the first to incorporate human activity maps for functional realism, but it struggled with misclassifications and a lack of user interaction. 
% Shabani \etal~\cite{shabani2023housediffusion} introduced HouseDiffusion, a diffusion-based approach that denoises room and door corner coordinates, utilizing transformer attention masks to integrate input graph constraints and directly generate vectorized floor plans.
HouseDiffusion~\cite{shabani2023housediffusion} generated vectorized floor plans by denoising room coordinates through a transformer-based diffusion model with graph constraints.
%Dupty \etal~\cite{dupty2024constrained} used factor graphs to model pairwise and higher-order dependencies among room bounding boxes and spatial attributes. % remove this?
Cons2Plan~\cite{hong2024cons2plan} improved flexibility by predicting connected edges to handle multiple conditional inputs, enhancing diversity in layout generation. 
% Hu \etal~\cite{hu2025gsdiff} proposed GSDiff, a model for generating wall junctions and predicting wall segments, with an emphasis on maintaining geometric consistency through structured graphs.
GSDiff~\cite{hu2025gsdiff} generated wall junctions and segments while maintaining geometric consistency through structured graphs.
%Despite the advantages of graph-based methods, they face challenges in controllability, where the generated layouts may appear random and struggle with consistency under complex user constraints.
%Our approach addresses this gap by supporting both apartment-level constraint management and controllable room-level editing, offering a distinct advantage in floorplan design.
Graph-based methods are effective when reliable topology constraints are available, but they often require explicit graph inputs or post-processing, and are less direct for localized editing when users provide sparse or evolving constraints.

\subsection{Text-constrained Generation}

Recent advances in multi-modal models have introduced natural language as a conditioning signal for floor plan generation~\cite{chen2020intelligent}. 
Several works enable human-in-the-loop design by translating textual or sketch-based prompts into spatial constraints, allowing interactive layout generation and refinement~\cite{leng2023tell2design,qin2024chathousediffusion}. 
Other approaches leverage large language models (LLMs) to generate structured layout descriptions that guide subsequent geometric synthesis, such as HouseLLM~\cite{zong2024housellm} and FP-LLaMa~\cite{yin2025floorplan}. 
These methods demonstrate the potential of natural-language-driven design interfaces, but still struggle to achieve precise control over local geometry and maintain spatial coherence during layout generation.
\section{Method}

\subsection{Gaussian Room Embedding}
\label{sec:gre}

Following prior floor plan generation works~\cite{shabani2023housediffusion,chen2024polydiffuse}, we represent each room in the layout as a polygonal region. 
Each room \(i\) is represented as a polygon \(x_0^i = [v_1^i, \dots, v_N^i]\),
where \(v_j^i \in \mathbb{R}^2\) denotes the \(j\)-th vertex, ordered clockwise to form a closed shape.
This polygon representation provides a precise vector description of room geometry and serves as the final layout produced by our model.
Motivated by the representation challenges discussed in Sec.~\ref{sec:intro}, we introduce \textbf{Gaussian Room Embedding (GRE)}, a continuous probabilistic representation that provides a structured room-level latent description of polygonal geometry. 
Specifically, each room is associated with a two-dimensional Gaussian distribution $g_i = \mathcal{N}(\mu_i, \sigma_i^2 I)$, where the mean \(\mu_i \in \mathbb{R}^2\) represents the room centroid and  \(\sigma_i\) is a scalar room-level scale parameter. In this work, we use an isotropic Gaussian parameterization, where \(\sigma_i^2 I \in \mathbb{R}^{2 \times 2}\) is not a full covariance matrix but a compact approximation of the room's spatial extent. Intuitively, \(\mu_i\) determines the room 
\begin{wrapfigure}[13]{r}{0.48\textwidth}
\centering
\includegraphics[width=\linewidth]{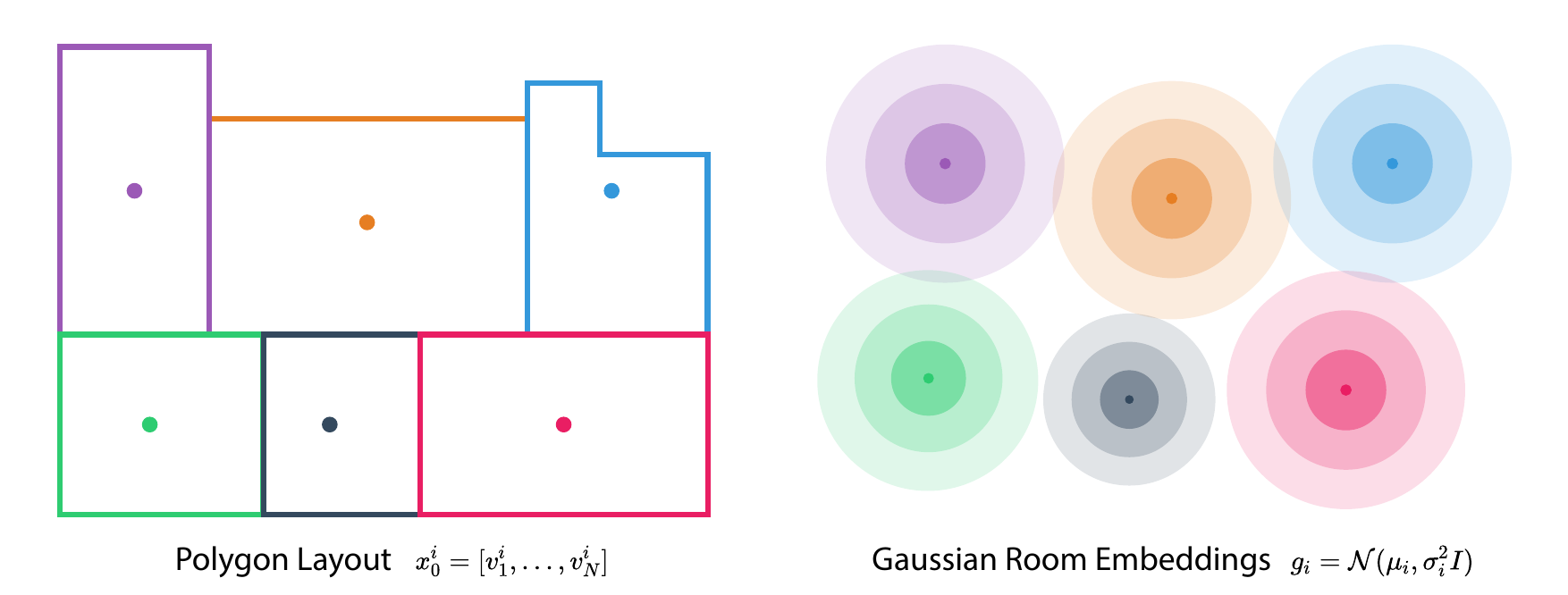}
\caption{
Polygon layout and its Gaussian Room Embedding.
GRE provides a continuous room-level spatial state that initializes and guides polygon diffusion, while the final output remains an explicit polygonal layout.
}
\label{fig:DM}
\end{wrapfigure}
location, while \(\sigma_i\) summarizes its coarse geometric scale and spatial influence within the layout. GRE does not replace the explicit polygon representation. Instead, it serves as a compact room-level spatial state that initializes and guides the diffusion process, reducing sensitivity to valid vertex-order variations and coordinate perturbations compared with directly diffusing ordered polygon vertices. The final output remains an explicit polygonal layout decoded by the denoising process.

\subsection{Gaussian-guided Diffusion Framework}

Building upon the Gaussian Room Embedding representation introduced in Sec.~\ref{sec:gre}, 
we formulate floor plan generation as a Gaussian-guided diffusion process. 
The proposed framework first predicts structured Gaussian room embeddings from user conditions and then generates polygonal room layouts through a conditional diffusion model.

\subsubsection{Diffusion Formulation.} Given the Gaussian room embeddings predicted by \emph{GuidanceNet}, 
we generate room polygons through a diffusion-based denoising process.
Instead of initializing the diffusion process from isotropic Gaussian noise, 
we sample the initial state of each room from its Gaussian embedding:
$x_T^i \sim \mathcal{N}(\mu_i, \sigma_i^2 I)$.
At each time step \( t \in [0, T]\), $\mu^i_t =(1-\sqrt{\overline\alpha_t})\mu_i$, $\sigma^i_t =(\sqrt{1-\overline\alpha_t})\sigma_i$.
The denoising process estimates the previous state \( x_{t-1}^i \) from the current noisy state \( x_t^i \) through the transition distribution \( p(x_{t-1}^i | x_t^i) \). 
The update rule is: 
\[ x_{t - 1}^i = \frac{1}{\sqrt{\alpha_t}} \left( x_t^i - \textcolor{orange}{\mu^i_t} - \frac{1 - \alpha_t}{1 - \overline{\alpha}_t} \textcolor{orange}{\sigma^i_t} \cdot \textbf {\textit D}(x_t, i, \textcolor{orange}{y}) \right) + \textcolor{orange}{\mu^i_t} + \sigma_t z^i ,\]
where $\alpha_t$ is the noise attenuation coefficient that controls the intensity of noise removal at each step, and \( z^i \sim \mathcal{N}(0, I) \) introduces stochasticity for sampling.
Here, $D(x_t, i, y)$ represents \emph{DenoisingNet}, whose architecture is described in the next section, and predicts the noise residuals conditioned on structural guidance \(y\), which includes  
(1) a \emph{boundary image} defining the apartment contour, and
% (2) \emph{semantic room labels} enforcing function-level consistency, and 
(2) \emph{anchor masks} indicating user-specified fixed regions during interactive editing.  
This conditional formulation enables the network to generate geometrically valid and semantically coherent layouts.

\subsubsection{GuidanceNet}

\begin{figure*}[!t]
    \centering
    \includegraphics[width=1.0\linewidth]{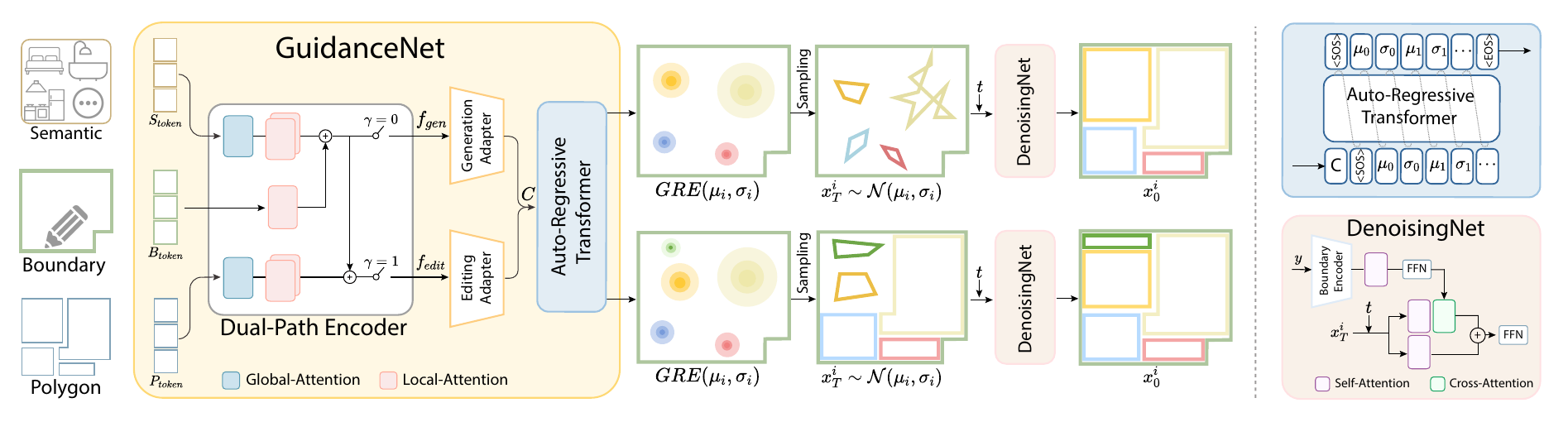}
    \caption{
    Inference process of \emph{GRE-Diff}.
\emph{GuidanceNet} encodes semantic ($S_{\text{token}}$), boundary ($B_{\text{token}}$), and polygon ($P_{\text{token}}$) conditions to predict Gaussian room embeddings $(\mu_i,\sigma_i)$. 
Samples drawn from $\mathcal{N}(\mu_i,\sigma_i^2 I)$ serve as diffusion initialization and are iteratively refined by \emph{DenoisingNet} under boundary constraints to generate vectorized layouts $x_0^i$. 
The resulting layouts can be re-encoded as polygon tokens, enabling iterative refinement and interactive editing.
    }
    \label{fig:network}
    %\vspace{-10pt}
\end{figure*}

GuidanceNet consists of four modules:
a multi-conditional encoder that tokenizes user inputs,
a dual-path encoder for semantic--geometric feature fusion,
task-specific adapters for generation and editing modes,
and an autoregressive transformer that sequentially predicts GRE.

\textit{Multi-conditional Encoding.}
As illustrated in Fig.~\ref{fig:network}, GuidanceNet encodes multi-modal user conditions into structured Gaussian representations that capture room position, size, and type.
It transforms input conditions into three types of tokens: room semantics ($S_{token}$), apartment boundaries ($B_{token}$), and geometric polygons ($P_{token}$), respectively.
$S_{token}$ encodes functional intent derived from the LLM-parsed JSON that specifies room types and quantities (\eg, three bedrooms and one kitchen).
$B_{token}$ describes the apartment contour using 64 uniformly sampled points, providing a global geometric constraint that bounds the layout.
$P_{token}$ represents existing room geometries with vertex-level precision and acts as a spatial prior for refinement.

\textit{Dual-path Encoder.}
As illustrated in Fig.~\ref{fig:network}, the dual-path encoder jointly processes semantic, boundary, and polygonal conditions through complementary attention pathways.
Specifically, $S_{token}$ and $P_{token}$ are independently processed by separate global attention modules, each modeling long-range spatial dependencies within its respective token domain.
No cross-attention is applied between these branches, allowing functional semantics and geometric priors to be learned independently before feature fusion.
Meanwhile, boundary tokens $B_{token}$ are processed by a local attention branch that focuses on fine-grained geometric alignment along the apartment contour.
The overall encoded feature representation is formulated as:

\[
C = \mathrm{Adapter}\bigl(
\mathrm{Enc}(S_{token}) +
\mathrm{Enc}(B_{token}) +
\gamma \mathrm{Enc}(P_{token})
\bigr),
\]
where $\mathrm{Enc}(\cdot)$ denotes the corresponding attention-based encoder for each token type, and $\mathrm{Adapter}(\cdot)$ denotes the task-specific adapter for either generation or editing.
The coefficient $\gamma$ modulates the contribution of polygonal cues, with $\gamma=0$ for generation and $\gamma=1$ for editing.
This design enables GuidanceNet to maintain a unified semantic--geometric latent space while adapting its representation to layout generation and editing.

\textit{Autoregressive Transformer.}
To sequentially generate room representations conditioned on semantic and geometric context, we employ an autoregressive transformer.
At each step, the transformer predicts the Gaussian parameters of the $i$-th room based on the previously generated ones:
\[
(\mu_i, \sigma_i)
=
\mathrm{ARTrans}
\bigl(
C,\,
\langle \mathrm{SOS} \rangle,\,
\langle \mu_{<i}, \sigma_{<i} \rangle
\bigr),
\]
where $\langle \mu_{<i}, \sigma_{<i} \rangle$ denotes the sequence of previously predicted parameters, and $C$ provides global semantic--geometric conditioning.
Starting from $\langle \mathrm{SOS} \rangle$, the model auto-regressively outputs $(\mu_0, \sigma_0), (\mu_1, \sigma_1), \ldots, (\mu_N, \sigma_N)$, where each pair $(\mu_i, \sigma_i)$ defines the spatial center and extent of the $i$-th room.
This formulation models spatial dependencies among rooms, including adjacency, functional grouping, and relative scaling, thereby preserving coherent layout structure at both global and local levels.

\subsubsection{DenoisingNet}

DenoisingNet serves as the geometric decoder of \emph{GRE-Diff}, transforming Gaussian embeddings from GuidanceNet into explicit polygonal room layouts through a conditional diffusion process.
As shown in Fig.~\ref{fig:network}, DenoisingNet employs a dual-attention transformer tailored for vectorized floor plan representation and refinement.
A boundary encoder~\cite{he2016resnet} extracts geometric constraints and fuses them with latent room embeddings through cross-attention, while a parallel self-attention pathway captures inter-room spatial dependencies.
The fused features are then processed through feed-forward layers to predict $D(x_t, i, y)$, guiding the denoising trajectory toward structurally consistent configurations.
During editing, additional anchor masks within $y$ enable localized resampling and refinement while maintaining the global topology.

\subsection{User-guided Layout Generation and Editing} \label{sec:overview}

\begin{figure}[t]
  \centering
  \includegraphics[width=0.9\linewidth]{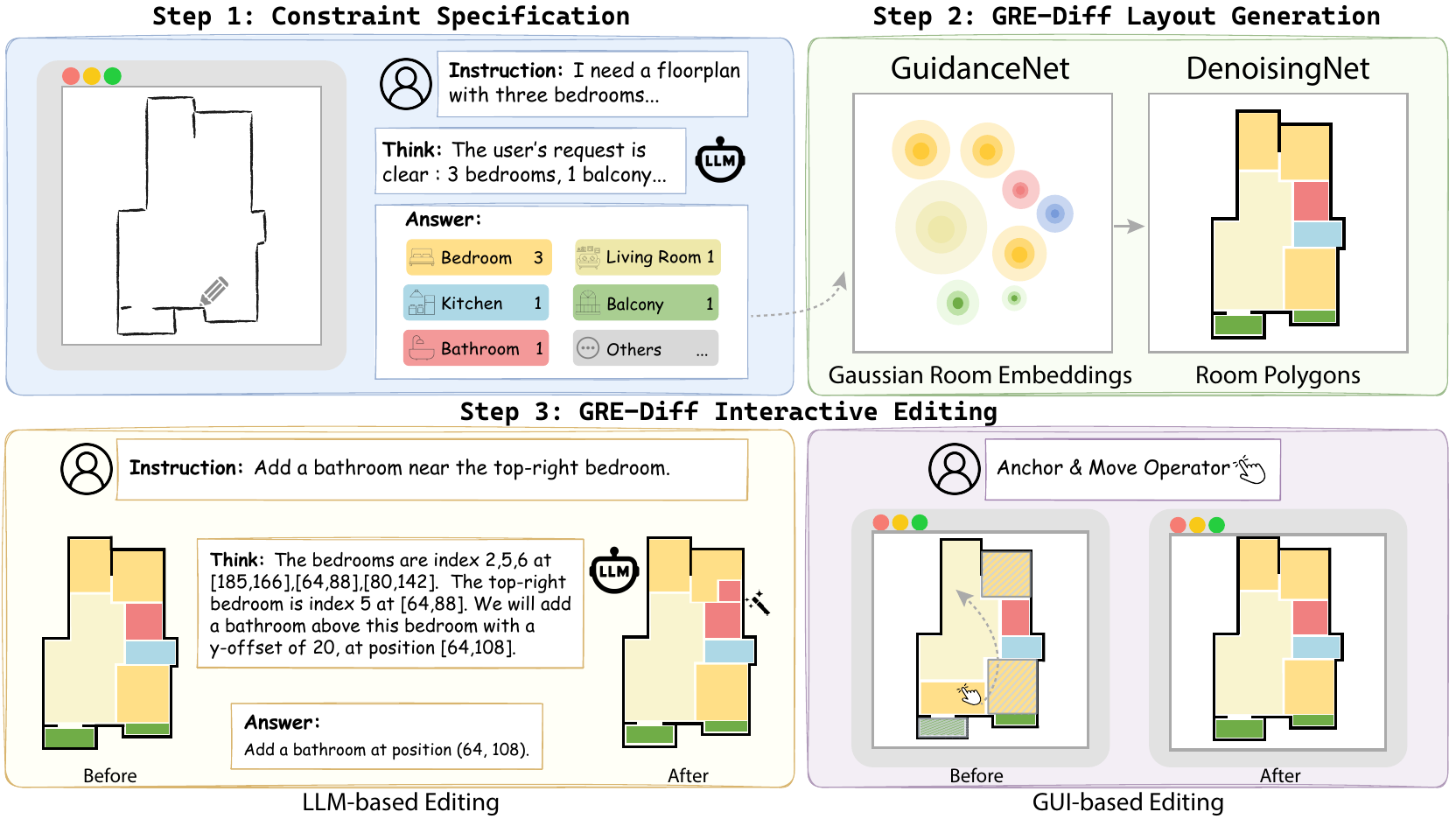}
  \caption{
Overview of \emph{GRE-Diff}, a controllable framework for floor plan generation and editing.
Given apartment boundaries and room constraints, \emph{GuidanceNet} predicts Gaussian room embeddings that guide the diffusion process, while \emph{DenoisingNet} generates coherent polygonal layouts.
The framework supports both language-based and GUI-based editing for interactive layout refinement.
}
  \label{fig:teaser}
  %\vspace{-10pt}
\end{figure}

As illustrated in Fig.~\ref{fig:teaser}, \emph{GRE-Diff} consists of three stages:
constraint specification, layout generation, and dual-mode editing.
%The framework encodes multi-modal user inputs into Gaussian-guided representations and enables controllable and interactive floor plan generation within a unified diffusion process.

\noindent\textbf{Step 1: Initial Constraint Specification.} 
Users describe layout requirements using natural language (e.g., ``three bedrooms and one kitchen''). 
A large language model (LLM), such as Kimi-K2~\cite{team2025kimi}, converts the request into a structured layout specification that defines room categories and quantities.
Users can also provide apartment boundaries by sketching or uploading floor outlines, which determine the spatial extent of the layout. 
Compared with prior methods~\cite{hu2020Graph2Plan,zhang2024MaskPLAN} that rely on manually defined topological graphs or attributes, our interface enables more flexible and intuitive control through natural-language instructions. 
Additional details are provided in the supplementary material.

\noindent\textbf{Step 2: Layout Generation.}   
% The structured conditions are processed by \emph{GRE-Diff}, which integrates two core modules: \emph{GuidanceNet} and \emph{DenoisingNet}.  
% \emph{GuidanceNet} encodes the multi-modal tokens into Gaussian embeddings that capture spatial and semantic priors,  
% and \emph{DenoisingNet} refines them through conditional diffusion to produce coherent polygonal layouts that satisfy user constraints.
These structured conditions are subsequently processed by \emph{GRE-Diff}, which integrates two core modules: \emph{GuidanceNet} and \emph{DenoisingNet}. Specifically, \emph{GuidanceNet} encodes the multi-modal tokens into Gaussian embeddings that capture both spatial and semantic priors, while \emph{DenoisingNet} iteratively refines them through conditional diffusion to produce coherent polygonal layouts that strictly satisfy user constraints.

\noindent\textbf{Step 3: Dual-Mode Editing.}  
To refine a generated layout, \emph{GRE-Diff} supports interactive editing through either LLM-based instructions or polygon-level manipulation via a GUI. The framework provides five basic editing operators: \textit{Add}, \textit{Delete}, \textit{Move}, \textit{Repurpose}, and \textit{Anchor}. The first four operators modify room attributes such as location or type, whereas \textit{Anchor} fixes selected rooms during editing.
\emph{GRE-Diff} does not regenerate the entire layout from scratch. Instead, user-anchored rooms are enforced as hard constraints through a fixed-room index mask: after each denoising step, their polygon vertices are copied back from the input layout. The remaining editable rooms are resampled under the predicted GRE priors, the boundary condition (y), and the current layout context. This mechanism enables localized refinement while preserving fixed structures and maintaining global spatial coherence.

\section{Results and Evaluation}
\subsection{Experimental Setup}

\subsubsection{Dataset.}

We use the RPLAN dataset~\cite{wu2019Data-Driven}, which contains over 80,000 vectorized floor plans with detailed room types, boundaries, and structural annotations. Its diverse residential layouts reflect real-world complexity and are widely adopted in prior works~\cite{wu2019Data-Driven, hu2020Graph2Plan}. For our experiments, we select six room types: living room, bedroom, kitchen, bathroom, dining room, and balcony. The data is split into 72,709 samples for training and 8,079 for testing.

\subsubsection{Metrics.}
Following prior works~\cite{hu2025gsdiff,hong2024cons2plan,gao2022COVMMD}, we evaluate our method across seven dimensions:
\textit{Distribution similarity} (FID~\cite{heusel2017gansFID}, 
KID~\cite{binkowski2018KID}, MMD~\cite{achlioptas2018learningCOVMMDProposed,gao2022COVMMD}), 
\textit{Generative diversity} (COV~\cite{gao2022COVMMD}), 
\textit{Generative novelty} (LPIPS~\cite{zhang2018LPIPS}), 
visual realism, 
\textit{Conditional controllability} (BC, RC, $F1_{BC\text{-}RC}$), 
\textit{Editing Effectiveness} (Tiny-ROE~\cite{ye2025TINY}), and \textit{Efficiency} (parameters, inference time). 
Metrics are averaged over five runs using 512 randomly sampled test layouts.
For \textit{functional validity}, the overlap-free rate measures the percentage of generated layouts without positive-area intersections between any pair of room polygons. The reachability rate evaluates whether the room connectivity graph forms a single connected component, where rooms are treated as nodes and derived door connections are treated as edges. These metrics complement BC and RC by assessing whether the generated layouts are not only constraint-satisfying but also spatially functional.
Additional details are provided in the supplementary materials.

%Further details are provided in the supplementary materials.
%\textit{Distribution similarity} is measured using FID, KID, and MMD to quantify the alignment between generated and real distributions. 
%\textit{Generative diversity} is assessed by COV and the variance of LPIPS, capturing variability under identical conditions and mitigating mode collapse. 
%\textit{Generative novelty} is evaluated with LPIPS to ensure the generated layouts differ from the training samples. 
%\textit{Visual realism} is evaluated via a user study measuring perceptual authenticity. 
%\textit{Conditional controllability} is quantified by Boundary Constrain (BC) and Room Constrain (RC), reflecting adherence to boundary- and room-level conditions. 
%\textit{Editing effectiveness} is measured by LPIPS for perceptual consistency and constraint-based metrics for geometric and semantic preservation after editing. 

\subsubsection{Training.}
\emph{GRE-Diff} is trained for 300 epochs on eight NVIDIA Quadro GV100 GPUs with a batch size of 40 per GPU. 
We use the AdamW optimizer~\cite{loshchilov2017adamw} with a learning rate of $2\times10^{-4}$, applying linear warm-up followed by cosine decay. 
Further details of the forward and reverse diffusion processes and loss formulations are provided in the supplementary materials.

\subsection{Evaluation of Generation Capability}

\subsubsection{Quantitative Evaluation with SOTAs.}

We evaluate the generation performance of \emph{GRE-Diff} against seven state-of-the-art methods, including iPLAN~\cite{he2022iplan}, Graph2Plan~\cite{hu2020Graph2Plan}, HouseDiffusion~\cite{shabani2023housediffusion}, WallPLAN~\cite{sun2022WallPlans}, MaskPLAN~\cite{zhang2024MaskPLAN}, GSDiff~\cite{hu2025gsdiff}, and ChatHouseDiffusion~\cite{qin2024chathousediffusion}.
All models are evaluated under identical constraint conditions to ensure fair comparison.
HouseDiffusion is evaluated without boundary constraints, as its original formulation does not support explicit boundary conditioning. 
Integrating boundary constraints would require non-trivial architectural modifications beyond the scope of the original model. 
%Metrics cover \textit{Similarity} (FID, KID, MMD), \textit{Diversity} (COV), \textit{Controllability} (BC, RC, $F1_{BC\text{-}RC}$), and \textit{Efficiency} (parameters, inference time).

% As summarized in Table~\ref{tab:evolution},
%Table~\ref{tab:evolution} shows that \emph{GRE-Diff} consistently outperforms all baselines. 
%It achieves the lowest FID of 4.36, outperforming GSDiff, Graph2Plan, ChatHouseDiffusion, and HouseDiffusion by 4.2\%, 9.4\%, 16.3\%, and 33.4\%, respectively. 
%It also records the lowest KID ($0.96\times10^{-3}$) and MMD (0.107), indicating strong geometric consistency and visual fidelity. 
%In terms of \textit{Diversity}, our model attains the highest coverage (96.09\%), showing its ability to explore broader layout variations. 
%For \textit{Controllability}, it achieves BC = 98.44\% and RC = 100.00\%, resulting in an $F1_{BC\text{-}RC}$ of 99.21\%, the best among all baselines.
%Despite having only 60M parameters, \emph{GRE-Diff} achieves superior results and generates one layout in 0.91s, faster than iPLAN and ChatHouseDiffusion, and comparable to WallPLAN. 

Table~\ref{tab:evolution} shows that \emph{GRE-Diff} achieves the best overall performance among the compared methods. 
It obtains the lowest FID of 4.36, improving over GSDiff, Graph2Plan, ChatHouseDiffusion, and HouseDiffusion by 4.2\%, 9.4\%, 16.3\%, and 33.4\%, respectively. 
It also achieves the lowest KID ($0.96\times10^{-3}$) and MMD (0.107), indicating strong distributional similarity to real layouts. 
In terms of generative diversity, \emph{GRE-Diff} attains the highest coverage of 96.09\%, showing its ability to explore diverse layout variations. For controllability, it achieves a high BC of 98.44\% and the best RC of 100.00\%, resulting in the highest $F1_{BC\text{-}RC}$ score of 99.21\%. 
With 60M parameters, \emph{GRE-Diff} generates one layout in 0.91s, faster than iPLAN and ChatHouseDiffusion and comparable to WallPLAN.

We further note that Graph2Plan relies on a retrieval-assisted refinement pipeline to obtain its final layouts. 
As a diagnostic comparison, removing this non-neural refinement increases its FID to 20.10 and reduces RC to 50.19\%, with BC and $F1_{BC\text{-}RC}$ dropping to 0.00\%. This suggests that its final performance depends heavily on post-hoc refinement. 
In contrast, \emph{GRE-Diff} directly synthesizes vectorized polygonal layouts within a unified diffusion framework, while also supporting local editing under user-specified constraints.

\newcolumntype{L}[1]{>{\raggedright\arraybackslash}p{#1}}
\begin{table}[t]
\centering
\belowrulesep=0pt
\aboverulesep=0pt
\caption{Comparison with prior floor plan generation and editing methods. A checkmark (\checkmark) indicates the available conditioning type (boundary, bubble, room type). The best cells are highlighted in light green and the second best in light blue. Our method attains the best overall similarity scores and the highest controllability with competitive efficiency among diffusion-based models.}
\label{tab:evolution}
\setlength{\tabcolsep}{3.8pt}
\renewcommand{\arraystretch}{1.15}
\rowcolors{4}{gray!3}{white}
\resizebox{\textwidth}{!}{
\begin{tabular}{
>{\centering\arraybackslash}p{2.9cm}|c c c |c c c | c | c c c |c c}
\toprule
% \multirow{2}{*}{\textbf{Network}} &
\multirow{2}{*}{\textbf{Method}} &
\multirow{2}{*}{\textbf{Bound.}} &
\multirow{2}{*}{\textbf{Bubble}} &
\multirow{2}{*}{\textbf{RType}} &
% \multirow{2}{*}{\textbf{Output}} &
\multicolumn{3}{c}{\textbf{Similarity (}\,$\downarrow$\,\textbf{)}} &
\multicolumn{1}{c}{\textbf{Diversity (}\,$\uparrow$\,\textbf{)}} &
% \textbf{Diversity (}\,$\uparrow$\,\textbf{)} &
\multicolumn{3}{c}{\textbf{Controllability (}\,$\uparrow$\,\textbf{)}} &
\multicolumn{2}{c}{\textbf{Efficiency (}\,$\downarrow$\,\textbf{)}} \\
\cmidrule(lr){5-13}
  &  &  &  & 
FID & KID$_{(\times10^{-3})}$ & MMD &
COV &
BC & RC & $F1_{BC\text{-}RC}$ &
Param (M) & Time (S) \\
\bottomrule

Graph2Plan~\cite{hu2020Graph2Plan}        & $\checkmark$ &  $\checkmark$  & $\checkmark$  &
4.81 & 1.65 & 0.124 &
\secondup{94.92\%} &
\bestup{100.00\%} & 83.11\% & \secondup{90.78\%} &
\bestdown{8}  & \bestdown{0.41} \\

iPLAN~\cite{he2022iplan}             & $\checkmark$ &            & $\checkmark$  &
12.63 & 12.20 & 0.148 &
75.11\% &
47.53\% & 85.65\% & 60.75\% &
46  & 1.61 \\

 HouseDiffusion~\cite{shabani2023housediffusion}    &              & $\checkmark$ & $\checkmark$   &
6.55 & 2.83 & 0.193 &
76.36\%  &
-- & \secondup{93.35\%} & -- &
\seconddown{27} & 8.76 \\

 WallPLAN~\cite{sun2022WallPlans}          & $\checkmark$ &            &               &
6.73 & 4.46 & 0.149 &
58.40\% &
99.01\% & -- & -- &
106 & 0.72 \\

 MaskPLAN~\cite{zhang2024MaskPLAN}          & $\checkmark$ &            & $\checkmark$  &
23.54 & 22.84 & 0.190 &
67.18\% &
\bestup{100.00\%} & 20.12\% & 33.50\% &
998 & 2.68 \\

 GSDiff~\cite{hu2025gsdiff}            & $\checkmark$ &            &               &
\seconddown{4.55} & \seconddown{1.13} & \seconddown{0.118} &
86.13\% &
91.61\% & -- & -- &
137 & \seconddown{0.67} \\

 ChatHouseDiff.~\cite{qin2024chathousediffusion}    & $\checkmark$ & $\checkmark$  & $\checkmark$  &
5.21 & 1.31 & 0.145 &
83.01\% &
99.80\% & 53.91\% & 70.01\% &
82 & 1.15 \\

 \textbf{Ours}     & $\checkmark$ &            &               &
\bestdown{\textbf{4.36}} & \bestdown{\textbf{0.96}} & \bestdown{\textbf{0.107}} &
\bestup{\textbf{96.09\%}} &
98.44\% & \bestup{\textbf{100.00\%}} & \bestup{\textbf{99.21\%}} &
60 & 0.91 \\

\bottomrule
\end{tabular}
}
%\vspace{-10pt}
%\raggedright\footnotesize
%\textit{Notes:} “--” denotes not reported / not applicable and is excluded from best/second ranking. 
%Best cells are light green; second-best are light blue. Arrows indicate the preferred direction.
\end{table}

\begin{figure}[!t]
\includegraphics[width=0.88\linewidth]{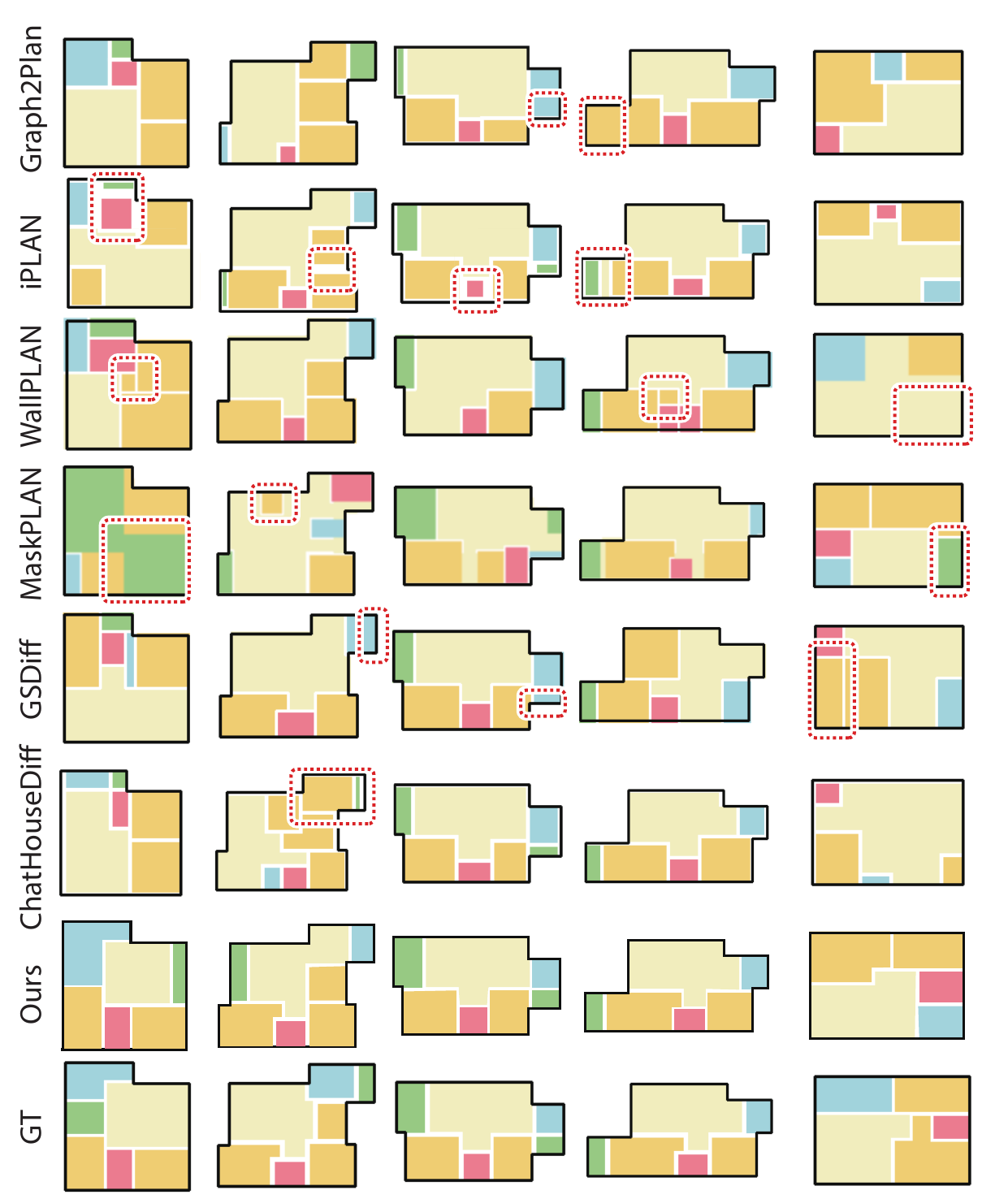}
\centering
\caption{Qualitative comparison of floor plan generation.
The red dashed regions highlight failure cases in existing methods, often arising from inconsistencies in room numbers, functional mismatches, or inaccessible layouts.
\emph{GRE-Diff} generates structurally coherent and functionally consistent results, closely resembling GT layouts.}
%In contrast, Graph2Plan occasionally omits rooms, HouseDiffusion introduces unintended holes in the layout, and MaskPLAN produces blurred room boundaries, reducing spatial clarity. }
%\vspace{-10pt}
\label{fig:test3}
\end{figure}

Existing methods exhibit different trade-offs between geometric validity, room-level controllability, diversity, and post-processing dependency.
{Graph2Plan} depends on retrieval and boundary alignment, achieving high controllability but low diversity.
{iPLAN} and {MaskPLAN} use progressive room localization and attribute decomposition with heavy post-processing, often producing over-dispersed layouts or empty regions. 
{WallPLAN} and {GSDiff} generate wall-line polygons via snapping and alignment, improving local precision but reducing layout diversity and failing to meet room-level constraints.
{HouseDiffusion} and {ChatHouseDiffusion} omit post-processing entirely, resulting in disconnected or inconsistent room geometries.
In contrast, \emph{GRE-Diff} embeds both boundary and room constraints directly into the diffusion dynamics, ensuring structural correctness, maintaining high diversity, and achieving real-time efficiency within a compact model.

\subsubsection{Qualitative Evaluation with SOTAs.}
To better illustrate the quantitative superiority, Fig.~\ref{fig:test3} visualizes representative results from different methods. 
Graph2Plan and WallPLAN often fail to generate all required rooms, resulting in incomplete layouts.
iPLAN and ChatHouseDiffusion produce unintended gaps that break spatial continuity and hinder accessibility.
MaskPLAN frequently exhibits discontinuous or fragmented boundaries, further reducing geometric consistency.
In contrast, \emph{GRE-Diff} preserves complete room structures, produces smooth and closed boundaries, and maintains balanced spatial organization, consistent with its strong numerical performance. More qualitative results are presented in the supplementary materials.

\begin{figure*}[t]
\includegraphics[width=0.9\linewidth]{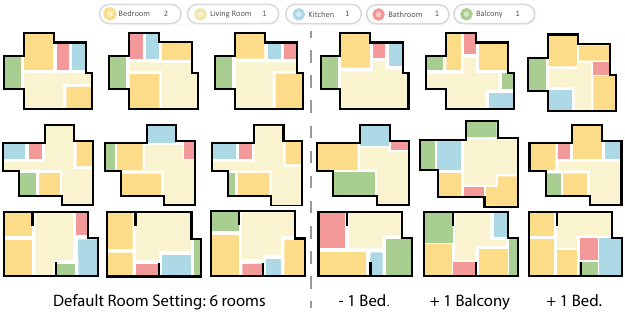}
\centering
\caption{Results of floor plans generated under various configurations. This figure shows the network's adaptability in generating layouts with consistent room configurations and its flexibility in adjusting room arrangements within a fixed boundary, underscoring the impact of load-bearing wall constraints. The network autonomously optimizes room layouts while maintaining spatial organization and functionality.}
\label{fig:test_default}
%\vspace{-10pt}
\end{figure*}

\subsubsection{Diversity.}
As illustrated in Fig.~\ref{fig:test_default}, we examine the controllable diversity of \emph{GRE-Diff} under various input conditions. 
(1) With identical functional requirements (\eg, one living room, one balcony, one bathroom, and two bedrooms), the model generates structurally diverse yet functionally consistent layouts, demonstrating its stochastic generative capability (the first three columns). 
(2) When room types vary while the building boundary remains fixed, the model adaptively reorganizes the layout to preserve functional logic, \eg, aligning balconies with exterior walls and placing living rooms at the center (last three columns).
(3) Under structural constraints such as fixed load-bearing walls, our method naturally respects these limitations and maintains spatial organization (the last row). 
These results collectively demonstrate that \emph{GRE-Diff} achieves both high generative diversity and controllable structural consistency.

\subsubsection{Realism.} 
We conducted a user study to evaluate the quality of our generated floor plans against Graph2Plan and ground truth (GT) layouts. Using boundaries, room types, and counts extracted from GT as constraints, we assessed 30 floor plans divided into two groups: one without room location input (compared to GT), and one with location input (compared to Graph2Plan). Participants, unaware of layout sources, selected their preferred design from randomized options.
We invited 58 computer science graduate students, collecting 1,740 responses. Preference counts were: GT/Ours/Cannot tell — 342/479/49; Graph2Plan/Ours/Cannot tell — 242/568/60. These results indicate that participants preferred our layouts over GT in 60.69\% of cases, and over Graph2Plan in 70.12\%, demonstrating the strong realism of our generated floor plans. 

\subsubsection{Novelty.} 
\begin{wrapfigure}[15]{r}{0.5\textwidth}
% \vspace{-6pt}
% \begin{wrapfigure}[15]{r}[0em]{0.5\textwidth}
% \centering
\includegraphics[width=\linewidth]{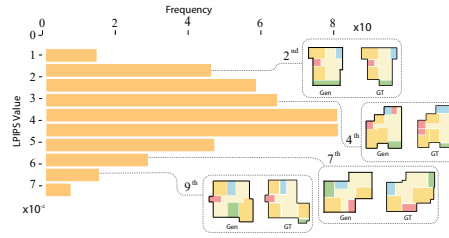}
\caption{Novelty analysis using LPIPS. 
We plot the LPIPS distribution of 500 randomly generated layouts and visualize representative samples from different percentile groups together with their nearest retrieved training counterparts. }
%The results show that GRE-Diff produces layouts that are both realistic (low LPIPS) and structurally diverse (high LPIPS).}

\label{fig:test_novelty}
%\vspace{-6pt}
\end{wrapfigure}

We assess perceptual similarity between generated and reference layouts using LPIPS~\cite{zhang2018LPIPS}.
As shown in Fig.~\ref{fig:test_novelty}, most scores lie within the range 
[0.03, 0.05], indicating high perceptual alignment, while samples above 0.06 correspond to intentionally diverse geometric configurations. 
For illustration, we display samples from the 2nd, 4th, 7th, and 9th groups along with their most similar counterparts retrieved from the training set. 
Visualizations across LPIPS ranges highlight both close resemblance (low LPIPS) and novelty (high LPIPS), demonstrating the model's ability to generate realistic and diverse layouts.
These results confirm that \emph{GRE-Diff} preserves structural fidelity while enabling diverse layout variations.

% \subsubsection{Functionality.}
% We further evaluate the functional usability of generated layouts for downstream applications. Although the generated results provide semantic room labels and vectorized geometric representations, practical use also requires basic geometric validity and spatial connectivity. Therefore, we add two quantitative checks on the same evaluation set, covering 19,680 generated layouts. For overlap, a layout is counted as valid only when no pair of room polygons has an invalid intersection; 95.56\% of layouts are overlap-free. For reachability, following prior rule-based floor plan post-processing methods~\cite{wu2019Data-Driven}, we derive door connections from shared boundaries between adjacent rooms and build a room connectivity graph, where rooms are nodes and door connections are edges. A layout is considered reachable only when this graph contains a single connected component. Under this criterion, 96.35\% of layouts are reachable. These results show that \emph{GRE-Diff} produces layouts with strong geometric validity and basic functional connectivity, making them suitable for downstream tasks such as furniture arrangement.    

\subsubsection{Functionality.}
We further evaluate the functional usability of generated layouts for downstream applications by measuring overlap-free rate and reachability rate on 19,680 generated layouts. A layout is considered overlap-free if no pair of room polygons has a positive-area invalid intersection; 95.56\% of the generated layouts satisfy this criterion. For reachability, we use the annotated door labels to construct a room connectivity graph, where rooms are nodes and door connections are edges. A layout is considered reachable if all rooms belong to a single connected component through the labeled door connections. Under this criterion, 96.35\% of the generated layouts are reachable. These results indicate that \emph{GRE-Diff} produces largely non-overlapping and door-connected layouts, supporting downstream applications such as furniture arrangement and interior layout refinement.

\subsection{Evaluation of Editing Capability}

\subsubsection{Quantitative Evaluation with SOTAs. }
As shown in Table~\ref{tab:edit_compare}, our method achieves a $0.00\%$ Tiny-ROE across all operations, indicating that every requested edit is correctly applied. 
In terms of editing accuracy, our F1 scores reach $86.27\%$ for \textit{Add}, $98.58\%$ for \textit{delete}, and $97.59\%$ for \textit{Anchor \& Move}, 
which outperform the best baseline (ChatHouseDiffusion) by $+17.8$, $+40.3$, and $+24.5$ points, respectively. 
%Failure rates are also drastically reduced, from $67.1\%$, $58.9\%$, and $46.6\%$ to $23.3\%$, $1.3\%$, and $4.1\%$. 
Although MaskPLAN achieves a $0.00\%$ Tiny-ROE, it suffers from much lower F1. 
These results demonstrate that our model not only executes all edit operations successfully but also maintains superior geometric and semantic consistency across diverse editing types.

\begin{table}[h]
\centering

\begin{minipage}{0.46\textwidth}
\centering
\caption{Editing performance comparison across different operation types (\textit{Add}, \textit{Delete}, and \textit{Anchor \& Move}).}
\label{tab:edit_compare}
\setlength{\tabcolsep}{4pt}
\renewcommand{\arraystretch}{1.2}
\resizebox{\textwidth}{!}{
\begin{tabular}{ccccccc}
\toprule
\multirow{2}{*}{\textbf{Method}} &
\multicolumn{2}{c}{\textbf{Add}} &
\multicolumn{2}{c}{\textbf{Delete}} &
\multicolumn{2}{c}{\textbf{Anchor \& Move}} \\
\cmidrule(lr){2-7}
 & Tiny-ROE (\%) $\downarrow$ & $F1_{BC\text{-}RC}$ $\uparrow$ &
Tiny-ROE (\%) $\downarrow$ & $F1_{BC\text{-}RC}$ $\uparrow$ &
Tiny-ROE (\%) $\downarrow$ & $F1_{BC\text{-}RC}$ $\uparrow$ \\
\midrule
MaskPLAN &
\bestup{0.00} & 49.47 &
\bestup{0.00} & 24.09 &
\bestup{0.00} & 66.05 \\

ChatHouseDiffusion &
68.49 & 68.45 &
10.96 & 58.26 &
4.11 & 73.06 \\

\textbf{Ours} &
\bestup{0.00} & \bestup{86.27} &
\bestup{0.00} & \bestup{98.58} &
\bestup{0.00} & \bestup{97.59} \\
\bottomrule
\end{tabular}}
\end{minipage}
\hfill
\begin{minipage}{0.48\textwidth}
\centering
\caption{Ablation study on the effect of \textit{Global Attention} (\emph{GA}) and \textit{Local Attention} (\emph{LA}) in \emph{GuidanceNet}.}
\label{tab:ablation_attention}
\resizebox{\textwidth}{!}{
\begin{tabular}{c c c c c c c c c}
\toprule
\textbf{GA} & \textbf{LA} & \textbf{FID}$\downarrow$ & \textbf{KID$_{(\times10^{-3})}$}$\downarrow$ & \textbf{MMD}$\downarrow$ & \textbf{COV}$\uparrow$ & \textbf{BC}$\uparrow$ & \textbf{RC}$\uparrow$ & {$\textbf{F1}_{BC\text{-}RC}$}$\uparrow$ \\
\midrule
\xmark & \cmark & 4.57 & 0.87 & 0.153 & 86.52\% & 96.67\% & 99.80\% & 98.21\% \\
\cmark & \xmark & 4.49 & \bestup{0.82} & 0.145 & 86.71\% & 97.07\% & 99.41\% & 98.23\% \\
\cmark & \cmark &
\bestup{4.36} &
0.96 &
\bestup{0.107} &
\bestup{96.09\%} &
\bestup{98.44\%} &
\bestup{100.00\%} &
\bestup{99.21\%} \\
\bottomrule
\end{tabular}}

\end{minipage}

\end{table}

\begin{figure}[t]
\centering
\includegraphics[width=0.9\linewidth]{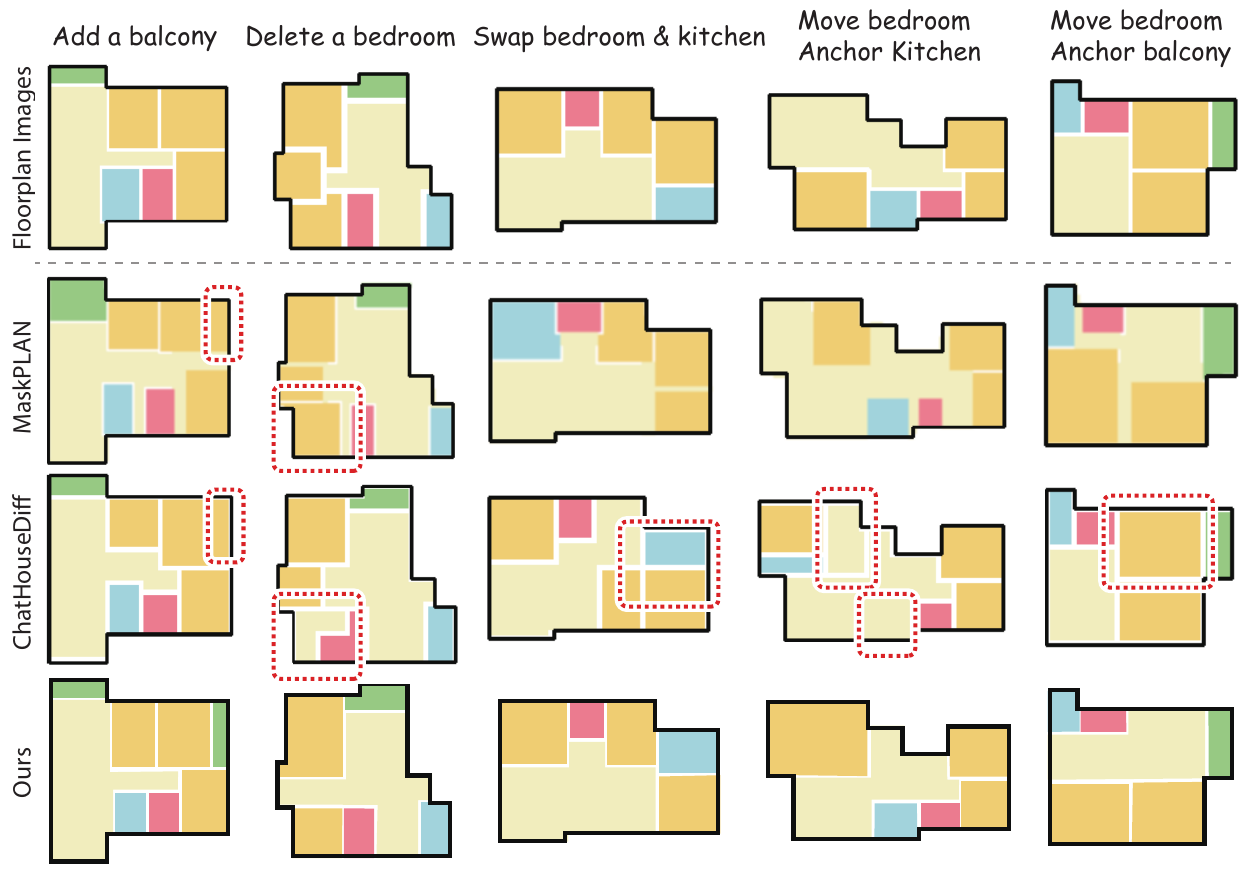}
\caption{Qualitative comparison of floor plan editing. We apply editing operations to the original ground truth layout, as shown in the first row.  Compared to other methods, our approach provides more precise control over room positions, ensuring that fixed structures (balcony or kitchen) remain unchanged while enabling flexible room relocation. Red dashed boxes mark incorrect edits from other methods. }

\label{fig:test_editing}
\centering
%\vspace{-10pt}
\end{figure}
\begin{figure}[t]
\centering
\includegraphics[width=0.9\linewidth]{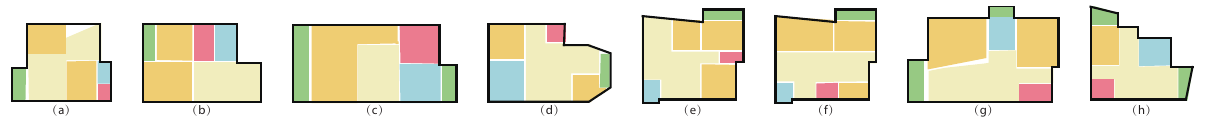}
\caption{Representative limitation cases.
Examples include (a) uneven vertex density; (b) incorrect room connectivity; (c) semantic mismatch; (d) limited generalization to non-rectilinear boundaries; (e-f) diverse layouts under identical non-Manhattan boundary; (g) a tilted bedroom; and (h) correctly places two balconies along a sloped wall.
Despite these, the layouts remain structurally coherent and spatially plausible. }
\label{fig:failure}
%\vspace{-10pt}
\end{figure}

\subsubsection{Qualitative Evaluation with SOTAs.}

Fig.~\ref{fig:test_editing} shows our model achieving superior editing consistency and spatial preservation.
It maintains fixed structures (\eg, balcony or anchored rooms) while adaptively updating surrounding regions. 
In contrast, MaskPLAN often suffers from boundary drift and unassigned spaces caused by mask misalignment.
We additionally implemented the \textit{Move} and \textit{Anchor} operations in ChatHouseDiffusion for a fair comparison; however, the method frequently produces unstable results with fragmented or overlapping rooms.
Additional qualitative results and implementation details are provided in the supplementary material.

\subsubsection{Automatic Refinement.} Beyond user-guided editing, \emph{GRE-Diff} integrates an automatic refinement mechanism based on BC and RC. This system provides real-time feedback to detect and correct issues such as overlaps, voids, or irregular geometries during generation. See the supplement for more details. 

\subsection{Ablation Studies}

Table~\ref{tab:ablation_attention} evaluates the roles of {global attention} (\emph{GA}) and {local attention} (\emph{LA}) in \emph{GuidanceNet}. 
Using only \emph{LA} weakens global guidance and reduces layout plausibility (FID = 4.57). 
Using only \emph{GA} better preserves global structure (FID = 4.49) but lacks room-level differentiation, leading to imbalanced room scales and lower COV (86.71\%). 
Combining \emph{GA} and \emph{LA} leverages their complementary strengths, achieving the best performance with the lowest FID (4.36) and highest COV (96.09\%). 
Additional ablations, including GRE-guided diffusion variants and robustness analyses, are provided in the supplementary material. 
Detailed GRE ablations in the supplementary material further show that integrating GRE into the diffusion dynamics is more effective than using GRE only as an external condition, validating its role as a room-level geometric state.
%Additional ablations on the effectiveness of \emph{GuidanceNet} are provided in the Supplementary Material. 

\subsection{Limitations Analysis}
While \emph{GRE-Diff} achieves strong overall performance, it occasionally fails under complex or incomplete spatial constraints (\eg, irregular boundaries), as shown in Fig.~\ref{fig:failure}.
In such cases, the model may produce suboptimal room adjacencies or disproportionate room sizes, especially when boundary geometries deviate markedly from the training distribution.
Examples (a–c) correspond to regular boundaries with uneven vertex density or misplaced functional zones, whereas (d–h) depict non-orthogonal or highly irregular contours that can lead to distorted room shapes.

\begin{figure}[t]
\includegraphics[width=\linewidth]{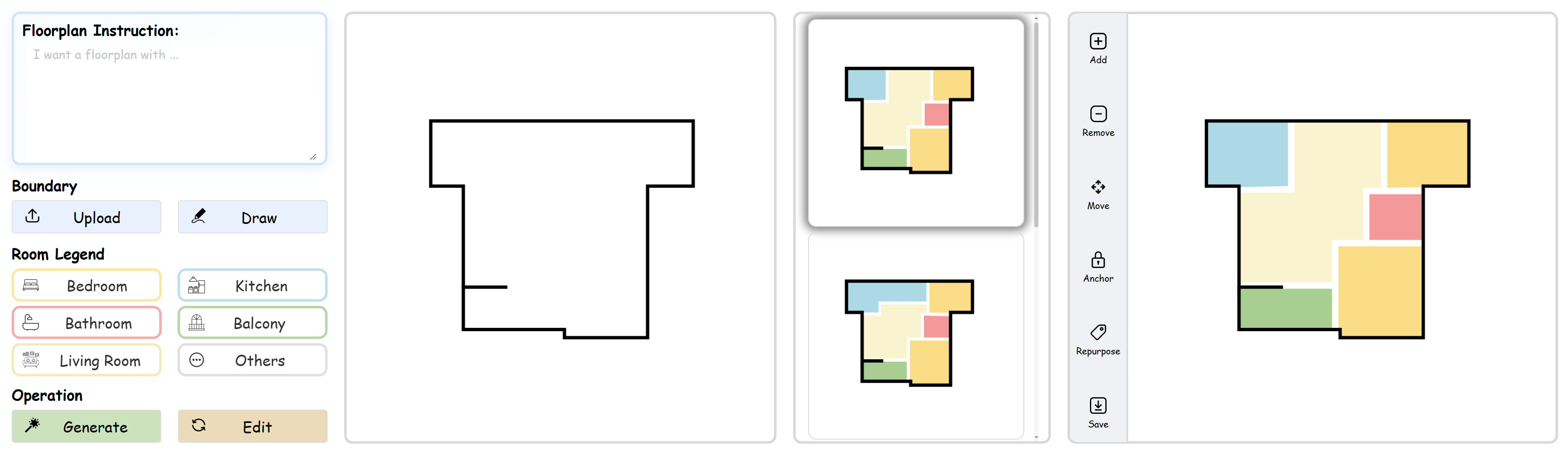}
\centering
%\vspace{-8pt}
\caption{
% Snapshot of the floorplan design interface, highlighting its intuitive and user-friendly experience.
Snapshot of the intuitive and user-friendly floor plan design interface.
}
\label{fig:demo}
%$\vspace{-6pt}
\end{figure}

\subsection{User Interface}

Fig.~\ref{fig:demo} shows the user interface of the floor plan design system.
Users can begin by providing natural-language instructions, while optionally uploading or sketching the apartment boundary (left panel).  
After clicking the \emph{Generate} button, \emph{GRE-Diff} produces multiple layout candidates that satisfy the given boundary and room-type constraints (center panel).  
Once a layout is selected, users can refine it interactively through intuitive operations such as \emph{Add}, \emph{Remove}, \emph{Move}, \emph{Anchor}, and \emph{Repurpose} (right panel).  
The refined floor plan can then be exported in vectorized form by clicking the \emph{Save} button.  
This interface provides a seamless and user-friendly workflow for iterative floor plan generation and editing.

\section{Conclusion and Future Work}
We presented \emph{GRE-Diff}, a controllable framework for floor plan generation and editing based on Gaussian Room Embeddings (GRE) and Gaussian-guided diffusion. 
The proposed design enables stable modeling of room configurations while supporting flexible user interaction.
Through its \emph{GuidanceNet} architecture, dual-path encoding, and real-time refinement workflow, \emph{GRE-Diff} produces high-quality, vectorized layouts that remain structurally coherent under diverse user constraints. By supporting both natural-language instructions and direct GUI manipulation, the system bridges automated layout synthesis with intuitive designer control, advancing the capabilities of AI-assisted spatial design. Future work will explore broader architectural typologies and extend the framework toward holistic indoor layout generation, including furniture placement for intelligent design automation.

%\textit{Limitations.} The current dataset primarily contains six-room layouts with Manhattan-structured walls, which limits \emph{GRE-Diff}'s ability to generalize to non-orthogonal or curved boundaries.This constraint may occasionally lead to minor inaccuracies in room connectivity or wall alignment.

\section*{Acknowledgments}
This work was supported in part by ICFCRT (W2441020), NSFC(62402323), Guangdong Basic and Applied Basic Research Foundation (2023B1515120026), Shenzhen S\&T Program (KQTD20210811090044003, KJZD20240903100022028), and Guangdong Provincial Key Laboratory of Visual Media and Multidimensional Intelligence.

\bibliographystyle{splncs04}
\bibliography{GRE-Diff}

\clearpage
\appendix
\section*{Supplementary Material}
\setcounter{figure}{0}
\renewcommand{\thefigure}{S\arabic{figure}}
\setcounter{table}{0}
\renewcommand{\thetable}{S\arabic{table}}
%\setcounter{page}{1}
%\maketitlesupplementary
%\appendix
\section{INTRODUCTION}
This supplementary document provides detailed information on various aspects of the \emph{GRE-Diff} paper, including the LLM prompt in Section~\ref{sec:LLM}, framework details in Section~\ref{sec:framework_details}, evaluation metrics in Section~\ref{sec:metrics}, and additional qualitative results in Section~\ref{sec:more_evaluations}.

\section{User-Guided Prompting with LLMs}
\label{sec:LLM}
Leveraging LLMs for natural-language-driven floor plan generation and editing reduces user effort and lowers the barrier for non-expert users. However, retraining or fine-tuning LLMs is impractical due to the high cost of data collection and the large number of parameters involved. Inspired by prior work~\cite{yao2025cast}, we adopt a few-shot prompting strategy that exploits the in-context learning ability of LLMs, enabling task learning from a small set of examples. This approach allows us to leverage large-scale LLM APIs for improved performance while ensuring consistent and structured output across various LLMs, making it scalable and efficient for a wide range of applications.

In the floor plan generation task, users can provide either detailed descriptions or abstract functional requirements (\eg, a compact apartment). The \textit{Question} represents the user’s instructions, \textit{Think} involves analyzing room types, functional relationships, and layout constraints, and \textit{Answer} generates the room types and quantities that fulfill the user’s needs.  This approach allows natural language interaction, which is more intuitive than traditional rule-based methods. Specific generation prompts can be found in Listing~\ref{lst:genere_prompt}.

For floor plan editing, the task is to convert user requirements into five basic operations. Starting with generated or user-uploaded vector polygons, we extract information such as center, area, and type in the \textit{Condition}. Given the user’s instruction in the \textit{Question}, the \textit{Think} analyzes spatial relationships, computes target values, and produces editing commands in the \textit{Answer} to perform the required edits. This two-step process enhances the system’s flexibility and adaptability, enabling real-time modifications to the floor plan. Detailed editing prompts are provided in Listing~\ref{lst:editing_prompt}.

% \lstset{
%   emph={Task Description, Relationship Definitions, Important Note, Output Format, Examples},
%   emphstyle=\bfseries\color{blue},
% }

\begin{lstlisting}[language={},
                   caption={Example of the generation prompt used by Kimi-K2},
                   label={lst:genere_prompt}]
(*@\textbf{\normalsize\textcolor{black}{Task Description}}@*)
You are a floor plan designer. Based on the floor plan description, you should analyze the room types and quantities. The known room types are living room, Bedroom, Kitchen, Bathroom, and Balcony. The total room count ranges from 4 to 8.



(*@\textbf{\normalsize\textcolor{black}{Examples}}@*)
(*@\textbf{\textcolor{black}{\textit{Question}}}:@*) I want a three-bedroom, one living room, one kitchen, one bathroom floor plan.
(*@\textbf{\textcolor{black}{\textit{Think}}}:@*) The user's request is clear: "three bedroom, one living room, one kitchen, one bathroom" means 3 bedrooms, 1 living room, 
1 kitchen, and 1 bathroom.
(*@\textbf{\textcolor{black}{\textit{Answer}}}:@*) {Livingroom: 1, Bedroom: 3, Kitchen: 1, Bathroom: 1, Balcony: 0}

(*@\textbf{\textcolor{black}{\textit{Question}}}:@*) I need a small apartment for person, with good natural light.
(*@\textbf{\textcolor{black}{\textit{Think}}}:@*) Since the user lives alone, only one bedroom is needed. A small apartment typically has no more than 5 rooms. Good lighting suggests the need for at least one balcony. Therefore, the required room types and quantities are: 1 living room, 1 bedroom, 1 kitchen, 1 bathroom, and 1 balcony.
(*@\textbf{\textcolor{black}{\textit{Answer}}}:@*) {Livingroom: 1, Bedroom: 1, Kitchen: 1, Bathroom: 1, Balcony: 1}

(*@\textbf{\textcolor{black}{\textit{Question}}}:@*) My spouse and I want a marital home, with spacious rooms.
(*@\textbf{\textcolor{black}{\textit{Think}}}:@*) The user is a couple, and it's a marital home. Besides the main bedroom, at least one additional bedroom is needed for guests or future children. Since the rooms should be spacious, we can consider 3 bedrooms and 2 bathrooms.
(*@\textbf{\textcolor{black}{\textit{Answer}}}:@*) {Livingroom: 1, Bedroom: 3, Kitchen: 1, Bathroom: 2, Balcony: 1}


(*@\textbf{\normalsize\textcolor{black}{User Input}}@*)
(*@\textbf{\textcolor{black}{\textit{Question}}}:@*) (*@\textbf{\textcolor{blue}{<Instruction>}}@*)
(*@\textbf{\textcolor{black}{\textit{Think}}}:@*)
(*@\textbf{\textcolor{black}{\textit{Answer}}}:@*)


\end{lstlisting}

% (*@\textbf{\textcolor{black}{\textit{Question}}}:@*) I want a floor plan with three bedrooms and one living room.
% (*@\textbf{\textcolor{black}{\textit{Think}}}:@*) The user's request is clear: three bedrooms and one living room. However, this does not follow typical floor plan standards, since a proper home must include at least one kitchen and one bathroom. Additionally, a layout with three bedrooms would be considered a relatively large apartment, so it should reasonably include a balcony as well. Therefore, it can be inferred that the user also needs one kitchen and one bathroom. The final room configuration is: 1 living room, 3 bedrooms, 1 kitchen, 1 bathroom, and 1 balcony.
% (*@\textbf{\textcolor{black}{\textit{Answer}}}:@*) {Livingroom: 1, Bedroom: 3, Kitchen: 1, Bathroom: 1, Balcony: 1}

% \lstset{
%   emph={Task Description, Relationship Definitions, Important Note, Output Format, Examples},
%   emphstyle=\bfseries\color{blue},
% }

\begin{lstlisting}[language={},
                   caption={Example of the editing prompt used by Kimi-K2},
                   label={lst:editing_prompt}]
(*@\textbf{\normalsize\textcolor{black}{Task Description}}@*)
You are a floor plan designer. Based on the users requirements, you need to analyze and generate the editing commands.
The coordinate system is a 2D Cartesian plane, with the origin at the top-left corner, the x-axis extending to the right, and the y-axis extending downward. so the bigger y value means lower position.
The known room types are: Living Room, Bedroom, Kitchen, Bathroom, and Balcony.
There are six types of editing commands:
1. Move room <index> to position (x,y)
2. Add a new <room_type> room at position (x,y)
3. Delete room <index>
4. Anchor room <index>
5. Change room <index> to <room_type>

(*@\textbf{\normalsize\textcolor{black}{Examples}}@*)
(*@\textbf{\textcolor{black}{\textit{Condition}}}:@*) [{index: 1, type: Living Room, area: 5394, center: [103, 120]}, {index: 2, type: Bedroom, area: 2684, center: [176, 89]}, {index: 3, type: Kitchen, area: 1178, center: [66, 86]}, {index: 4, type: Bathroom, area: 762, center: [125, 86]}, {index: 5, type: Bathroom, area: 400, center: [95, 80]}, {index: 6, type: Bedroom, area: 1560, center: [70, 169]}, {index: 7, type: Bedroom, area: 2193, center: [184, 142]}]
(*@\textbf{\textcolor{black}{\textit{Question}}}:@*) Move the bathroom to the right side of a bedroom.
(*@\textbf{\textcolor{black}{\textit{Think}}}:@*) The user wants to move a bathroom. There are two bathrooms, index 4 at [125,86] and index 5 at [95,80]. The bedrooms are index 2,6,7 at [176,89], [70,169], [184,142]. The bathroom to the left of a bedroom and closest is index 4 relative to bedroom index 2. We will move bathroom index 4 to the right side of bedroom index 2 with an x-offset. Consider the length is sqrt(area=2684) = 50, and Bathroom in the Bedroom. So, x-offset should be 50 / 2 = 25(if outside, 50). Bedroom center is [176,89], so the new bathroom position will be [201,89].
(*@\textbf{\textcolor{black}{\textit{Answer}}}:@*) [Move room 4 to position (201, 89)]

(*@\textbf{\textcolor{black}{\textit{Condition}}}:@*) [{index: 1, type: Living Room, area: 6140, center: [135, 110]}, {index: 2, type: Bedroom, area: 2519, center: [185, 166]}, {index: 3, type: Kitchen, area: 1041, center: [175, 87]}, {index: 4, type: Bathroom, area: 594, center: [198, 111]}, {index: 5, type: Bedroom, area: 1911, center: [64, 88]}, {index: 6, type: Bedroom, area: 1840, center: [80, 142]}, {index: 7, type: Balcony, area: 864, center: [133, 184]}]
(*@\textbf{\textcolor{black}{\textit{Question}}}:@*) Add a balcony to the top-left bedroom.
(*@\textbf{\textcolor{black}{\textit{Think}}}:@*) The bedrooms are index 2,5,6 at [185,166],[64,88],[80,142]. The top-left bedroom is index 5 at [64,88]. We will add a balcony above this bedroom with a y-offset of -20, at position [64,68].
(*@\textbf{\textcolor{black}{\textit{Answer}}}:@*) [Add a new Balcony room at position (64, 68)]

(*@\textbf{\textcolor{black}{\textit{Condition}}}:@*) [{index: 1, type: Living Room, area: 5864, center: [116, 118]}, {index: 2, type: Bedroom, area: 2620, center: [156, 121]}, {index: 3, type: Kitchen, area: 1000, center: [81, 51]}, {index: 4, type: Bathroom, area: 640, center: [169, 92]}, {index: 5, type: Bedroom, area: 2500, center: [160, 51]}, {index: 6, type: Bedroom, area: 1600, center: [113, 51]}, {index: 7, type: Balcony, area: 880, center: [81, 208]}, {index: 8, type: Balcony, area: 800, center: [160, 172]}]
(*@\textbf{\textcolor{black}{\textit{Question}}}:@*) The balcony near the bedroom is good, but the other one seems redundant.
(*@\textbf{\textcolor{black}{\textit{Think}}}:@*) The balconies are index 7 ([81,208]) and index 8 ([160,172]). Bedrooms are index 2,5,6 at [156,121],[160,51],[113,51]. The balcony closer to a bedroom is index 8, so we delete the other one, index 7, and anchor index 8 as the user is satisfied with it.
(*@\textbf{\textcolor{black}{\textit{Answer}}}:@*) [Delete room 7, Anchor room 8]

(*@\textbf{\normalsize\textcolor{black}{User Input}}@*)
(*@\textbf{\textcolor{black}{\textit{Condition}}}:@*) (*@\textbf{\textcolor{blue}{<Room Condition>}}@*)
(*@\textbf{\textcolor{black}{\textit{Question}}}:@*) (*@\textbf{\textcolor{blue}{<Instruction>}}@*)
(*@\textbf{\textcolor{black}{\textit{Think}}}:@*)
(*@\textbf{\textcolor{black}{\textit{Answer}}}:@*)


\end{lstlisting}

\section{FRAMEWORK DETAILS}
\label{sec:framework_details}

\subsection{Gaussian Room Embeddings}

Following prior diffusion-based floor plan generation works~\cite{shabani2023housediffusion,chen2024polydiffuse}, 
we represent each room as a polygon

\[
x_0^i = [v_1^i,\dots,v_N^i],
\]
where \(v_j^i\) denotes the 2D coordinates of the \(j\)-th vertex.

To describe the spatial configuration of rooms, we introduce 
\textbf{Gaussian Room Embeddings (GRE)} predicted by \emph{GuidanceNet}.
Each room \(i\) is associated with a Gaussian distribution

\[
g_i = (\mu_i, \sigma_i),
\]
which captures the approximate location and spatial extent of the room.
These embeddings provide structured spatial priors that guide the subsequent diffusion process.

\subsection{Gaussian-guided Diffusion}

Building upon GRE, we formulate layout generation as a \textbf{Gaussian-guided diffusion process}.  
Instead of initializing diffusion from a shared isotropic Gaussian distribution, each room evolves according to its own Gaussian embedding, allowing the diffusion trajectory to reflect room-specific spatial priors.

\paragraph{Forward Process}

For each room \(i\), the forward diffusion step is defined as

\begin{equation}
x_t^i =
\sqrt{\bar{\alpha}_t}\,x_0^i
+
\textcolor{orange}{\mu_t^i}
+
\textcolor{orange}{\sigma_t^i}z^i ,
\end{equation}
where \(z^i \sim \mathcal{N}(0,I)\).
The drift and variance schedules are derived from the Gaussian Room Embedding

\begin{equation}
\mu_t^i = (1-\sqrt{\bar{\alpha}_t})\mu_i,
\quad
\sigma_t^i = \sqrt{1-\bar{\alpha}_t}\sigma_i .
\end{equation}

This formulation introduces room-dependent drift and variance into the diffusion trajectory, 
where the mean anchors the room location and the scale controls its spatial extent.

\paragraph{Reverse Process}

During generation, layouts are reconstructed by iteratively denoising the latent variables.  
The reverse diffusion step is defined as

\begin{equation}
x_{t-1}^i =
\frac{1}{\sqrt{\alpha_t}}
\left(
x_t^i
-
\textcolor{orange}{\mu_t^i}
-
\frac{1-\alpha_t}{1-\bar{\alpha}_t}
\textcolor{orange}{\sigma_t^i}
\cdot
\mathbf{D}(x_t,i,y)
\right)
+
\textcolor{orange}{\mu_t^i}
+
\sigma_t z^i .
\end{equation}

Here \(\mathbf{D}(x_t,i,y)\) denotes the noise predicted by \emph{DenoisingNet}, and \(y\) represents structural conditions such as boundary constraints and optional anchor masks for interactive editing.
Although the update is applied independently to each room, \emph{DenoisingNet} processes all room polygons jointly.
This enables the predicted noise to capture global spatial relationships among rooms, ensuring both local geometric fidelity and global layout coherence.

\paragraph{Relation to Standard Diffusion}

Unlike conditioning-based methods, GRE directly parameterizes the diffusion state transition through 
\((\mu_t^i,\sigma_t^i)\).
Our formulation generalizes standard DDPM:
when \(\mu_i=0\) and \(\sigma_i=1\), the process reduces to conventional isotropic diffusion.

\subsection{GuidanceNet}

\emph{GuidanceNet} comprises two main components:  
(1) a \emph{Dual-Path Encoder} that extracts mid-level features and produces Gaussian Room Embeddings for each room; and  
(2) an \emph{Auto-Regressive Transformer} that autoregressively predicts room sequences conditioned on the encoder features, enabling diverse generation and controllable editing.  

\paragraph{Two-Stage Training Strategy.}
To stabilize optimization and prevent interference between feature learning and sequence generation, \emph{GuidanceNet} is trained in two successive stages.

\textbf{Stage 1: Encoder Training.}  
We first train the \emph{Dual-Path Encoder} using an auxiliary prediction head with the encoder loss \(\mathcal{L}_{\mathrm{enc}}\).  
The auxiliary head is implemented as a shallow MLP that predicts the Gaussian parameters \((\mu, \sigma)\) and room-level semantic labels.  
This explicit supervision encourages the encoder to learn robust intermediate features and well-structured room embeddings.  
After convergence, the \emph{Dual-Path Encoder} is frozen, and the auxiliary head is discarded.

\textbf{Stage 2: Autoregressive Training.}  
With the encoder fixed, we train the \emph{Auto-Regressive Transformer} using an autoregressive loss \(\mathcal{L}_{\mathrm{AR}}\).  
The transformer consumes the encoder-generated embeddings and sequentially predicts room tokens, supporting both diverse layout generation and controllable editings.

\paragraph{Dual-Path Encoder.}
The \emph{Dual-Path Encoder} is designed to extract robust mid-level features and generate meaningful
Gaussian embeddings \((\mu_i, \sigma_i)\) for each room.  
To ensure that these embeddings are geometrically stable, semantically informative, and invariant to polygon permutations, we introduce three complementary objectives:
(1) permutation-invariant discriminability,  
(2) semantic consistency, and  
(3) numerical stability.  
These are enforced through the permutation loss $\boldsymbol{\mathcal{L}_{\mathrm{perm}}}$, semantic loss $\boldsymbol{\mathcal{L}_{\mathrm{sem}}}$, and regularization loss $\boldsymbol{\mathcal{L}_{\mathrm{reg}}}$, respectively.

\textbf{1) Permutation Loss} $\boldsymbol{\mathcal{L}_{\mathrm{perm}}}$. A room embedding should be invariant to polygonal vertex permutations and remain consistent under forward-diffused variants.  
To achieve this, we adopt a bidirectional triplet loss that pulls embeddings of the same room closer while pushing embeddings of other rooms apart.  
For room \(i\), the triplet objectives are:

\begin{equation*}
\mathcal{L}_\mathrm{Tri.}(x_t^i, X_0) = 
\max \Big(
0, \, \max_{j \neq i } (\alpha +
\underbrace{D(x_t^i, x_0^i)}_{\text{positive}} -
\underbrace{D(x_t^i, x_0^j)}_{\text{negative}})
\Big), 
\vspace{-10pt}
\end{equation*}

\begin{equation*}
\mathcal{L}_\mathrm{Tri.}(x_0^i, X_t) = 
\max \Big(
0, \, \max_{j \neq i } (\alpha +
\underbrace{D(x_0^i, x_t^i)}_{\text{positive}} -
\underbrace{D(x_0^i, x_t^j)}_{\text{negative}})
\Big), 
\vspace{-10pt}
\end{equation*}

\begin{equation*}
    \mathcal{L}_{perm} = \sum_{i=1}^{N} (\mathcal{L}_\mathrm{Tri.}(x_t^i,X_0) + \mathcal{L}_\mathrm{Tri.}(x_0^i,X_t)), 
\end{equation*}
where  
\(X_0 = \{x_0^j\}_{j=1}^N\) denotes the set of GT polygons,  
\(X_t = \{x_t^j\}_{j=1}^N\) denotes their diffused variants at step \(t\),  
and \(\alpha = 0.1\) is a soft-margin hyperparameter.  
This loss enforces permutation invariance and robustness under forward diffusion.
During training, each \(x_0^i\) is passed through \emph{GuidanceNet} to obtain the room-level embeddings \((\mu_i,\sigma_i)\).  
These embeddings then parameterize the forward diffusion step to produce \(x_t^i\), ensuring that the triplet loss remains fully differentiable with respect to encoder parameters.

\textbf{2) Semantic Loss} $\boldsymbol{\mathcal{L}_{\mathrm{sem}}}$. Room embeddings should encode not only geometry but also semantics.  
To this end, we predict semantic labels directly from encoder features and supervise them using a Cross-Entropy loss:
\begin{equation*}
    \mathcal{L}_{sem} = \sum_{i=1}^{N} ( - \sum_{c=1}^{C} S_{gt}(i,c) \cdot log \frac{e^{S_{pred}(i,c)}} { \sum_{j=1}^{C} e^{S_{pred}(i,j)}} ),
\end{equation*}
where $N$ denotes the number of rooms, $C$ is the total number of semantic categories, $S_{gt}(i,c)$ is the one-hot ground-truth label for room $i$, and $S_{pred}(i,c)$ is the predicted logit for category $c$.

\textbf{3) Regularization Loss} $\boldsymbol{\mathcal{L}_{\mathrm{reg}}}$. To maintain numerical stability and avoid pathological variances, we regularize the predicted Gaussian parameters as:
\begin{equation*}
    \mathcal{L}_{reg} = \sum_{i=1}^{N}(||\mu_i||^2 + ||\frac{1}{\sigma_i}||^2).
\end{equation*}

The regularizer constrains the embeddings to remain numerically well-behaved, avoiding excessive mean offsets and preventing extremely small or large variances that may impair diffusion stability.

The final \emph{Dual-Path Encoder} objective, denoted as $\boldsymbol{\mathcal{L}_{\mathrm{enc}}}$, integrates the three loss components as:
\begin{equation*}
    \mathcal{L}_{enc} = \lambda_1 \mathcal{L}_{perm} + \lambda_2 \mathcal{L}_{sem} + \lambda_3 \mathcal{L}_{reg},
\end{equation*}
with weights \(\lambda_1 = 1.0\), \(\lambda_2 = 1.0\), and \(\lambda_3 = 0.5\).  
These constraints jointly produce stable and semantically expressive Gaussian Room Embeddings, enabling reliable room-level diffusion and controllable editing.

\paragraph{Auto-Regressive Transformer.}
Once robust room features and Gaussian Room Embeddings have been obtained, the \emph{Dual-Path Encoder} is frozen, and an \emph{Auto-Regressive Transformer} is trained to model room dependencies for unified floor plan generation and controllable editing.
Each Gaussian Room Embedding is parameterized by \((\mu_x, \mu_y, \sigma)\), which are discretized into \(K = 50\) bins through uniform quantization.  
Following a decoder-only autoregressive architecture, the Transformer predicts the token sequence for each room conditioned on previously generated tokens and the encoder features.  
This formulation enables the model to capture structural priors such as room ordering, adjacency relations, 
and functional co-occurrence patterns.
Training is performed using a Cross-Entropy objective over the quantized bins:
\begin{equation*}
\mathcal{L}_{\mathrm{AR}} =
\sum_{i=1}^{N}
\left(
    - \sum_{j=1}^{K}
    Q_{\mathrm{gt}}(i,j)
    \log
    \frac{
        e^{Q_{\mathrm{pred}}(i,j)}
    }{
        \sum_{k=1}^{K} e^{Q_{\mathrm{pred}}(i,k)}
    }
\right),
\end{equation*}
where \(Q_{\mathrm{gt}}(i,j)\) is the one-hot GT token for room \(i\) and bin \(j\), and 
\(Q_{\mathrm{pred}}(i,j)\) is the corresponding Transformer-predicted logit.
This autoregressive formulation allows the model to generate room sequences in a step-by-step manner and 
adaptively refine edited layouts, providing a unified mechanism for both free-form generation and 
fine-grained, controllable editing.

\subsection{DenoisingNet}

Given the Gaussian Room Embeddings, we sample the corresponding room polygons and refine them using
a room-level diffusion model.  
\emph{DenoisingNet} predicts noise residuals for each room and progressively denoises the polygons to
recover clean, vectorized floor plan geometry. 

The denoising objective is defined as:
\begin{equation*}
    \mathcal{L}_{\mathrm{denoise}}
    =
    \sum_{i=1}^{N}
    \left\lVert
        (x_t^i - x_0^i)
        -
        D(x_t^i, i, y)
    \right\rVert^2,
\end{equation*}
where \(x_t^i\) is the noisy polygon of room \(i\) at timestep \(t\),  \(x_0^i\) is its ground-truth polygon,  and \(D(\cdot)\) denotes the predicted noise residual.  
The conditioning input \(y\) includes all room-level contextual signals, such as the boundary map and the anchor-room mask, allowing the denoising process to incorporate both global boundary constraints and user-specified edits.
By predicting noise at the room level and conditioning on global layout context,  \emph{DenoisingNet} ensures that the recovered polygons remain geometrically coherent, functionally consistent, and aligned with user constraints.

\section{Evaluation Metrics}
\label{sec:metrics}

We comprehensively evaluate our method across four complementary dimensions, each aligned with prior
evaluation protocols adopted in recent generative modeling and conditional floorplan synthesis
research~\cite{hu2025gsdiff, hong2024cons2plan, zhang2024MaskPLAN, gao2022COVMMD, Ma2024COVMMD}.  
(1) \textbf{Distribution Similarity} measures how closely generated samples match the real
distribution, following widely used perceptual statistics such as FID and KID~\cite{heusel2017gansFID, binkowski2018KID}.  
(2) \textbf{Generative Diversity} evaluates variability and coverage of the real data modes, using
LPIPS-based COV/MMD metrics commonly employed in shape and layout generation~\cite{zhang2018LPIPS,
achlioptas2018learningCOVMMDProposed, Zhou2021COVMMD, YANG2019pointflowCOVMMD, Ma2025COVMMD2}.  
(3) \textbf{Conditional Controllability} assesses whether geometric and semantic constraints are
satisfied, following constraint-based evaluation established in Cons2Plan~\cite{hong2024cons2plan}
and MaskPLAN~\cite{zhang2024MaskPLAN}.  
(4) \textbf{Editing Effectiveness} examines whether user edits take effect and whether constraints
remain intact, building upon the Tiny-ROE family of editing metrics~\cite{ye2025TINY, zhang2023TINY,
brooks2023TINY}.  

For each evaluation, $|S_r|=512$ test floorplans are randomly sampled, and $|S_g|$ conditional
samples are generated. All results are averaged over five independent trials.  
Each floorplan is rendered as a $256\times256$ RGB image with a fixed color configuration and
4-pixel wall width to ensure consistent metric computation.

% ================================================================
% DISTRIBUTION SIMILARITY
% ================================================================
\subsection*{Fréchet Inception Distance (FID)}
Following~\cite{heusel2017gansFID}, FID computes the discrepancy between the Gaussian statistics of
generated and real samples extracted using an Inception-V3 encoder:
\begin{equation*}
\mathrm{FID}(S_g,S_r)
= \|m_g - m_r\|_2^2
+ \operatorname{Tr}(\Sigma_g + \Sigma_r - 2\sqrt{\Sigma_g \Sigma_r}),
\end{equation*}
where $(m_g,\Sigma_g)$ and $(m_r,\Sigma_r)$ denote means and covariances.  
We use torch-fidelity~\cite{obukhov2020torchfidelity} for implementation.

\subsection*{Kernel Inception Distance (KID)}
KID~\cite{binkowski2018KID} computes an unbiased squared MMD using a polynomial kernel $k(x,y)=\left(\frac{1}{d}x^\top y+1\right)^3$, offering
greater stability under small sample sizes:
\begin{equation*}
\begin{split}
\mathrm{KID}(S_g,S_r) &=
\frac{\sum_{i\ne j} k(x_i,x_j)}{|S_g|(|S_g|-1)}
+ \frac{\sum_{i\ne j} k(y_i,y_j)}{|S_r|(|S_r|-1)} \\
&- \frac{2}{|S_g||S_r|}
\sum_{i=1}^{|S_g|}\sum_{j=1}^{|S_r|}k(x_i,y_j).
\end{split}
\end{equation*}

% ================================================================
% GENERATIVE DIVERSITY
% ================================================================
\subsection*{Coverage (COV)}
To capture diversity and mode coverage, we adopt the LPIPS-based COV metric widely used in shape
generation literature~\cite{gao2022COVMMD, Ma2024COVMMD}. Following GET3D~\cite{gao2022COVMMD},
$|S_g|=5|S_r|$ samples are generated, and distances are computed using LPIPS~\cite{zhang2018LPIPS}.  
COV is defined as:
\begin{equation*}
\text{COV}(S_g,S_r)
=
\frac{1}{|S_r|}
\left|
\left\{
\arg\min_{x\in S_r} D(x,y) \;\middle|\; y\in S_g
\right\}
\right|.
\end{equation*}

% COV measures how well the generated samples cover the modes of the reference set, reflecting diversity and completeness of generation.

\subsection*{Minimum Matching Distance (MMD)}
MMD complements COV by emphasizing perceptual fidelity, as in~\cite{achlioptas2018learningCOVMMDProposed,
Ma2025COVMMD2, YANG2019pointflowCOVMMD}:
\begin{equation*}
\text{MMD}(S_g,S_r)
=
\frac{1}{|S_r|}
\sum_{x\in S_r}\min_{y\in S_g}D(x,y).
\end{equation*}

% It evaluates the proximity of each reference sample to the generated samples, thereby reflecting the perceptual similarity and overall quality of the generated set.

% ================================================================
% CONDITIONAL CONTROLLABILITY
% ================================================================
\subsection*{Boundary Constraint (BC)}
Following constraint-based floorplan evaluation in Cons2PLAN~\cite{hong2024cons2plan} and MaskPLAN
\cite{zhang2024MaskPLAN}, we define BC to check boundary validity and interior completeness:
\begin{equation*}
\text{BC}
=
\frac{1}{|S_g|}
\sum_{S_g}
(\mathbb{I}_{\mathrm{interior}}\land \mathbb{I}_{\mathrm{fill}})\times 100\%,
\vspace{-11pt}
\end{equation*}

\begin{equation*}
\mathbb{I}_{\mathrm{interior}}
=
\mathbb{I}\Big\{
\sum_{i,j}[M_r\land(1-M_b)]_{i,j}=0
\Big\},
\vspace{-11pt}
\end{equation*}

\begin{equation*}
\mathbb{I}_{\mathrm{fill}}
=
\mathbb{I}\Big\{
\sum_{u,v}[(1-M_r)\land M_b]_{i+u,j+v}<k^2
\Big\},
\end{equation*}
where $k=10$, $M_r$ is a binary mask whose value is $1$ inside room regions, and $M_b$ is a binary mask whose value is $1$ inside the boundary.

\subsection*{Room Constraint (RC)}
RC evaluates compliance with room-type and room-count specifications, also following
\cite{hong2024cons2plan, zhang2024MaskPLAN}:
\begin{equation*}
\text{RC}
=
\frac{1}{|S_g|}
\sum_{S_g}
\mathbb{I}\{\forall l:\; N_r(l)=N_c(l)\}\times 100\%,
\end{equation*}
where $l$ denotes a room type, $N_r(l)$ is the number of rooms of type $l$ in the generated floorplan, and $N_c(l)$ is the required count for type $l$.

\subsection*{F-measure (F1$_{BC\text{-}RC}$)}
This harmonic mean penalizes any imbalance in the satisfaction of geometric and semantic constraints.
\begin{equation*}
F1_{BC\text{-}RC}
=
2\cdot\frac{\text{BC}\cdot\text{RC}}{\text{BC}+\text{RC}}.
\end{equation*}

% ================================================================
% EDITING EFFECTIVENESS
% ================================================================
\subsection*{Tiny-ROE}
Tiny-ROE follows the formulation used in image editing benchmarks~\cite{ye2025TINY, zhang2023TINY,
brooks2023TINY}.  
We compute LPIPS between edited and original layouts and report the proportion with LPIPS $<0.02$,
indicating negligible modifications. Lower Tiny-ROE reflects more reliable edits.

 \section{MORE EVALUATIONS} \label{sec:more_evaluations}

% \input{tab/tabauto.tex}

% \begin{figure}[h]
%     \centering
%     \includegraphics[width=0.8\linewidth]{fig/Door.pdf}
%     \caption{Qualitative comparison of door and window placement strategies.}
%     \label{fig:test_Door}
%     \end{figure}

\begin{figure*}[ht]
    \centering
    \includegraphics[width=0.85\linewidth]{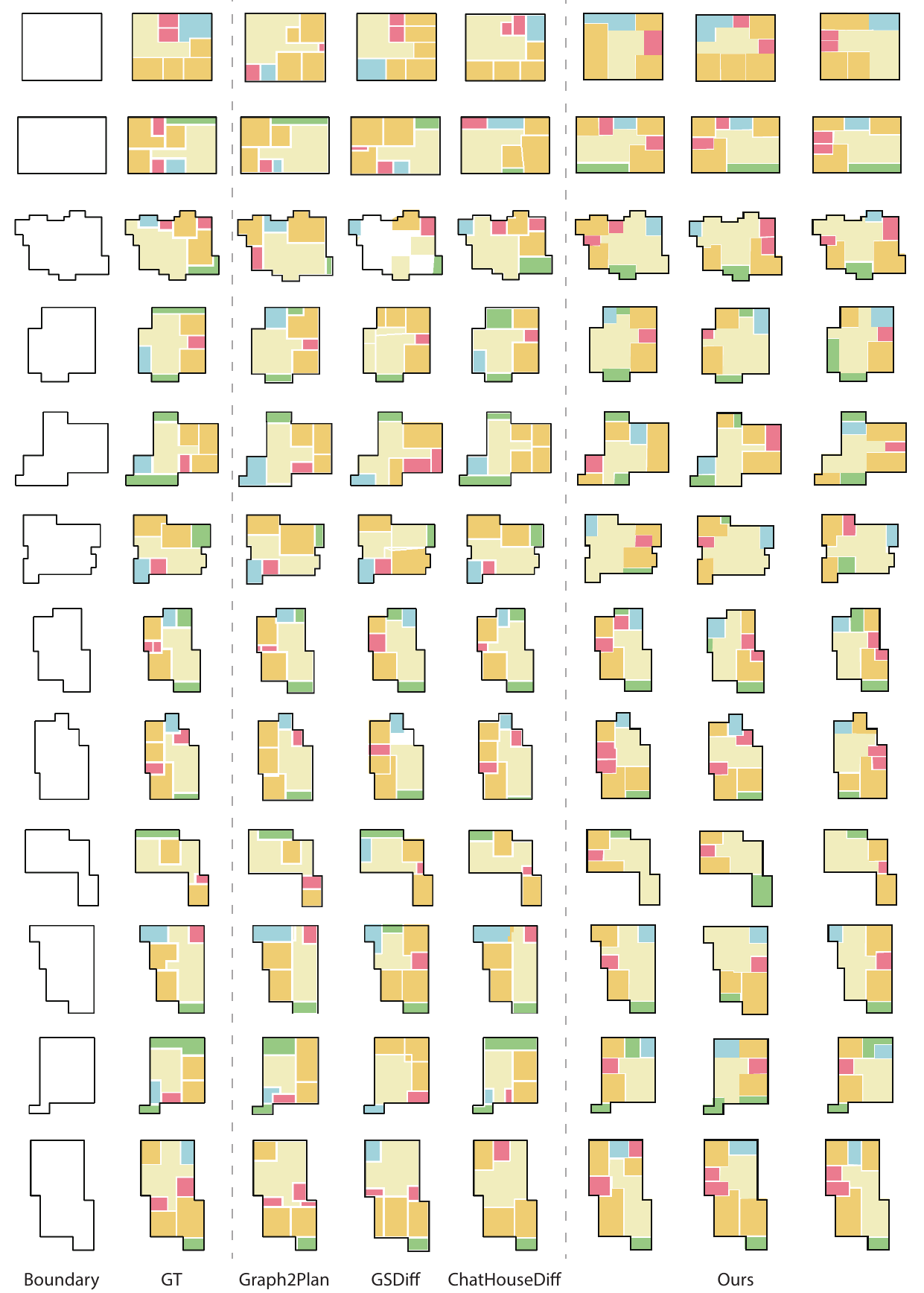}
    \caption{Comparison of floor plan generation with SOTA methods. The first column displays the boundary conditions, followed by the ground truth (GT) in the second column. The third to fifth columns show the results from Graph2Plan, GSDiff, and ChatHouseDiffusion, while the last three columns represent our method's results. This comparison demonstrates the accuracy and diversity of the generated floor plans, with our approach showing superior performance in both aspects, better capturing the structural and spatial features while maintaining high variability in the generated layouts.}
    \label{fig:test_sota}
    \end{figure*}

\begin{figure*}[ht]
    \centering
    \includegraphics[width=0.85\linewidth]{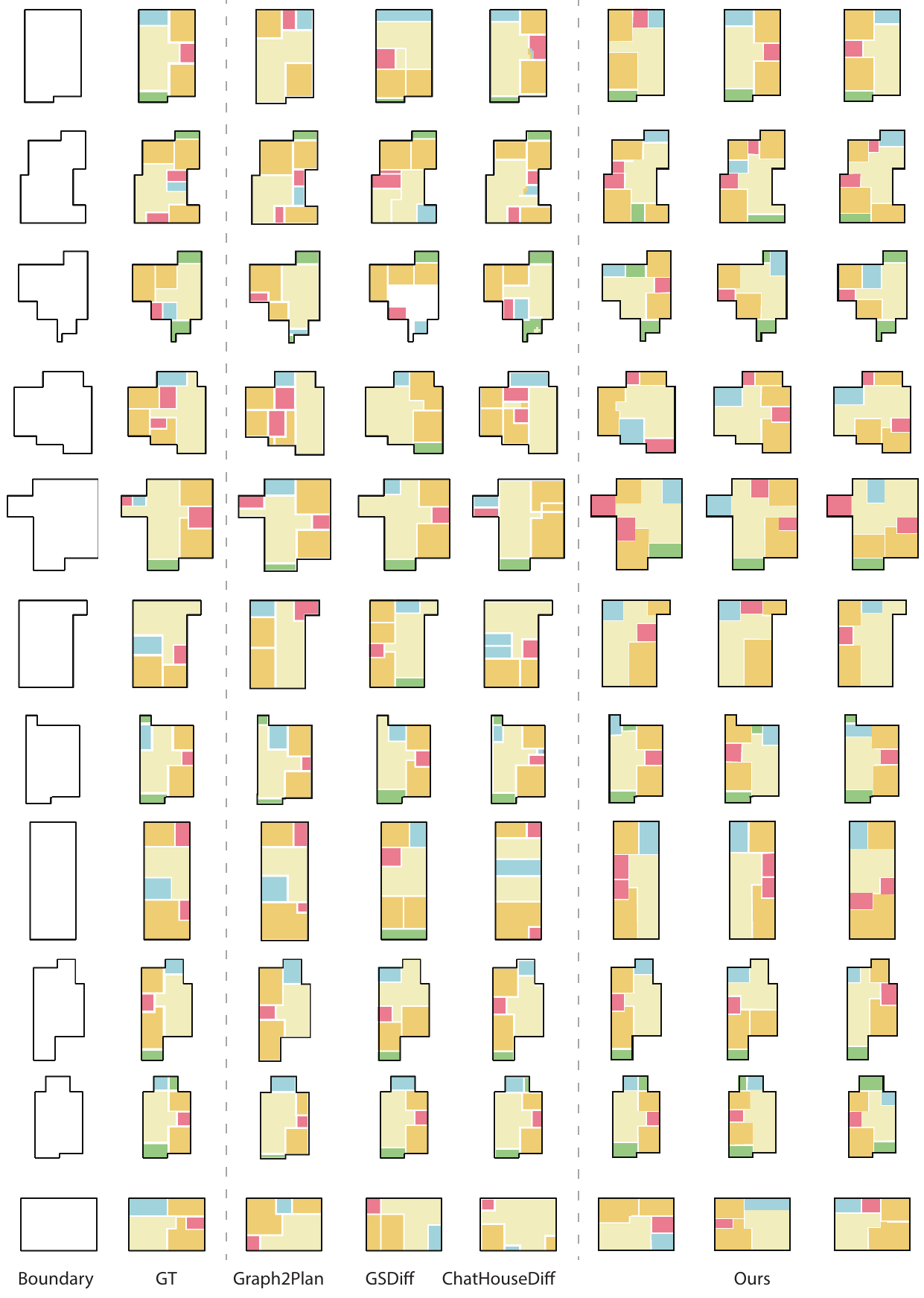}
    \caption{Comparison of floor plan generation with SOTA methods. The first column displays the boundary conditions, followed by the ground truth (GT) in the second column. The third to fifth columns show the results from Graph2Plan, GSDiff, and ChatHouseDiffusion, while the last three columns represent our method's results. This figure illustrates how our approach improves both accuracy and diversity of the generated floor plans compared to the other methods, with a stronger ability to maintain spatial coherence and adapt to different input conditions.}
    \label{fig:test_sota2}
    \end{figure*}

\paragraph{Qualitative Evaluation.}
The qualitative comparisons further highlight the advantages of \emph{GRE-Diff} in both floor plan generation and editing. As shown in Fig.~\ref{fig:test_sota} and Fig.~\ref{fig:test_sota2}, our method produces layouts that more accurately adhere to the boundary conditions while capturing the structural and spatial characteristics of real floor plans. Compared with state-of-the-art generative approaches such as Graph2Plan, GSDiff, and ChatHouseDiffusion, the proposed framework generates room arrangements that are geometrically coherent, functionally plausible, and noticeably more diverse. %In particular, our results better preserve overall apartment organization, correctly placing public and private zones and maintaining realistic room proportions even under complex boundary constraints.

In the editing comparison (Fig.~\ref{fig:test_editing}), \emph{GRE-Diff} demonstrates clear improvements in controllability and structural stability. When performing operations such as adding, removing, or relocating rooms, even under LLM-based user instructions, our method maintains consistent room adjacencies and avoids typical failure modes observed in prior works, such as broken connectivity, distorted shapes, or misaligned boundaries. Notably, when editing fixed or semantically important rooms (\eg, kitchens), our approach preserves global layout coherence while applying the requested modifications, whereas ChatHouseDiffusion often introduces geometric artifacts or inconsistencies. These examples collectively show that \emph{GRE-Diff} provides more reliable and structurally aware interaction, enabling precise user-guided refinements without compromising floor plan integrity.

\begin{figure}[!t]
\includegraphics[width=1\linewidth]{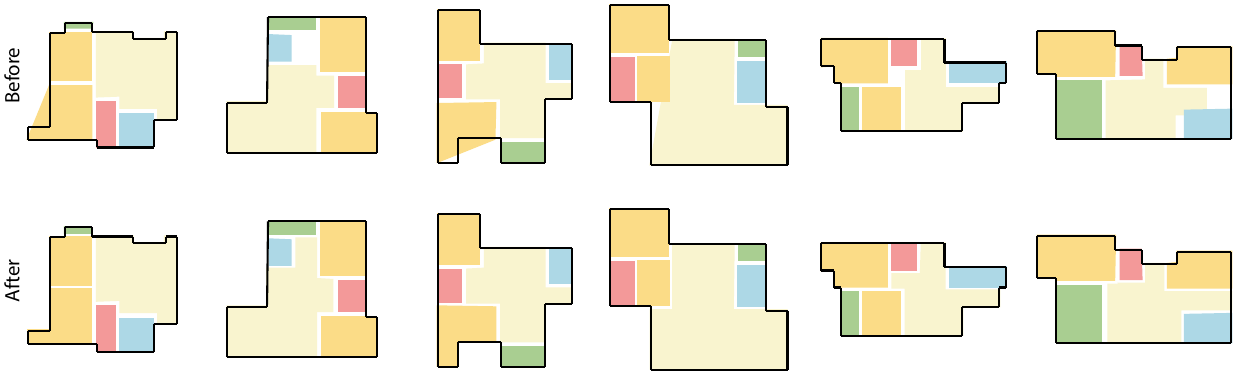}
\centering
\caption{Comparison of generated floorplan results before and after automatic optimization implementation.}
\label{fig:test_autorefine}

\end{figure}

\paragraph{Automatic Refinement.}
The constraint-based evaluation system built around \textit{Boundary Constraint} (BC) and \textit{Room Constraint} (RC) enables \emph{GRE-Diff} to perform automatic refinement of generated layouts. During inference, the model evaluates each sampled floor plan using BC and RC, and selectively regenerates layouts when these constraint scores fall below a predefined threshold. In this way, geometric and semantic violations, such as boundary leakage or room intersections, are corrected through constraint-guided resampling rather than heuristic post-processing, polygon snapping, or manual clean-up. 
This closed-loop mechanism allows \emph{GRE-Diff} to progressively improve layout validity, yielding floor plans that are structurally coherent and semantically compliant without any additional alignment or repair steps, as shown in Fig.~\ref{fig:test_autorefine}.

%\paragraph{Ablation Study.}
\begin{table}[t]
\centering
\caption{
Ablation and baseline comparison illustrating the role of Gaussian Room Embedding (GRE).
The comparison isolates whether GRE is used only as an external condition or integrated into the diffusion state transition.
}
\label{tab:ablation_attention}
\resizebox{0.9\columnwidth}{!}{
\begin{tabular}{c c c c c c c c}
\toprule
\textbf{Method} & \textbf{FID}$\downarrow$ & \textbf{KID$_{(\times10^{-3})}$}$\downarrow$ & \textbf{MMD}$\downarrow$ & \textbf{COV}$\uparrow$ & \textbf{BC}$\uparrow$ & \textbf{RC}$\uparrow$ & {$\textbf{F1}_{BC\text{-}RC}$}$\uparrow$ \\
\midrule
DDPM & 4.86 & 1.86 & 0.122 & 81.48\% & 81.64\% & 96.28\% & 88.39\% \\
DDPM + GRE (conditioning only) & 19.96 & 15.89 & 0.229 & 31.44\% & 9.96\% & 98.04\% & 18.08\% \\
Ours & {\bestup{4.36}} & {\bestup{0.96}} & {\bestup{0.107}} & {\bestup{96.09\%}} & {\bestup{98.44\%}} & {\bestup{100.00\%}} & {\bestup{99.21\%}} \\
\bottomrule
\end{tabular}}
\end{table}

\paragraph{Effect of GRE in the diffusion process.}
To isolate the contribution of Gaussian Room Embedding (GRE), we compare three configurations that differ only in how room-level Gaussian information is used during diffusion.
\textit{DDPM} denotes a standard diffusion baseline without GRE.
\textit{DDPM + GRE condition} predicts GRE with GuidanceNet but uses it only as an external conditioning signal for DenoisingNet.
\textit{GRE-Diff} denotes our full model, where GRE explicitly parameterizes the room-level diffusion state and directly participates in both initialization and denoising dynamics.
For a fair comparison, the conditioning-only variant uses the same network architecture and training settings as GRE-Diff, and is trained for 50\% longer.

As shown in Table~\ref{tab:ablation_attention}, the standard DDPM baseline achieves reasonable distributional similarity but shows limited controllability, especially in boundary compliance.
Adding GRE merely as an external condition does not improve the model; instead, it substantially degrades distribution similarity and boundary satisfaction, with FID increasing to 19.96 and BC dropping to 9.96\%.
This suggests that GRE is not effective when treated as auxiliary conditioning detached from the diffusion state.
In contrast, the full GRE-Diff formulation consistently achieves the best results across distribution similarity, diversity, and constraint satisfaction, including the lowest FID of 4.36, highest COV of 96.09\%, and highest $F1_{BC\text{-}RC}$ of 99.21\%.
These results indicate that the benefit of GRE comes from integrating it into the diffusion process as a room-level geometric state, rather than simply providing additional conditioning information.

\begin{figure*}[t]
\centering
\includegraphics[width=\textwidth]{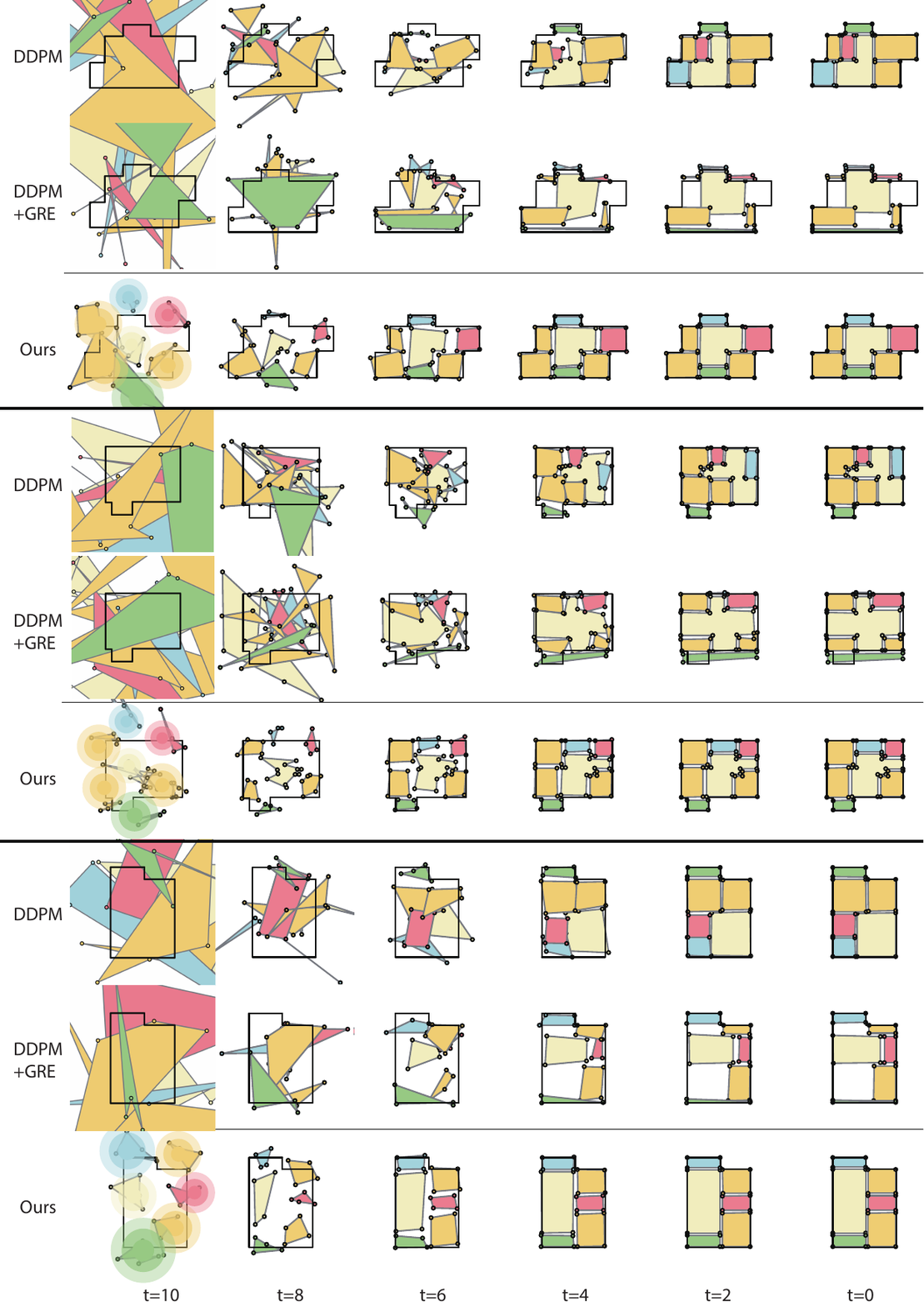}
\caption{
Diffusion trajectories under different configurations.
From top to bottom: DDPM, DDPM + GRE (conditioning), and GRE-Diff (ours).
Columns show diffusion steps from $t=10$ to $t=0$.
While DDPM and conditioning-based GRE exhibit unstable polygon evolution, GRE-Diff maintains structured trajectories and converges to coherent layouts.
}
\label{fig:diffusion_trajectory}
\end{figure*}

Fig.~\ref{fig:diffusion_trajectory} further illustrates the denoising trajectories of different variants.
The standard DDPM baseline can produce plausible final layouts, but its intermediate states evolve without explicit room-level spatial guidance, leading to unstable geometric trajectories.
The conditioning-only variant provides coarse room-location information, but the denoising process still struggles to maintain both spatial placement and polygonal regularity.
By contrast, GRE-Diff maintains structured room-level trajectories throughout denoising, enabling stable convergence toward coherent and geometrically regular layouts.
This confirms that GRE is most effective when it shapes the stochastic diffusion dynamics directly through Gaussian parameterization.

\paragraph{Robustness to valid perturbations.}
We further evaluate the stability of GRE under valid vertex-order variations and moderate coordinate perturbations.
For vertex-order variations, we apply cyclic shifts and reverse ordering to the vertices of each room polygon, which preserve the underlying polygonal geometry but change the vertex sequence.
For coordinate perturbations, we add small spatial noise to polygon vertices.
We report the changes in Gaussian mean $\Delta \mu$, Gaussian scale $\Delta \sigma$, and polygon-level deviation $\Delta P$ with respect to the original unperturbed input.

As shown in Table~\ref{tab:perturbation_analysis}, GRE remains stable under cyclic and reverse vertex-order variations.
The changes in Gaussian parameters are nearly zero, with $\Delta \mu=0.004$ and $\Delta \sigma=0.000$ for cyclic ordering, and $\Delta \mu=0.000$ and $\Delta \sigma=0.000$ for reverse ordering.
This indicates that GRE is less sensitive to valid vertex-order variations than direct polygon-coordinate representations.
Under moderate coordinate noise, GRE changes smoothly with the input geometry, showing controlled variations in $\mu$ and $\sigma$.
These results support the use of GRE as a compact room-level geometric state for structured diffusion.

\begin{table}[t]
\centering
\caption{
Robustness analysis under valid vertex-order variations and coordinate perturbations.
Smaller values indicate higher stability with respect to the original unperturbed input.
}
\label{tab:perturbation_analysis}
\begin{tabular}{lccc}
\toprule
\textbf{Perturbation} & $\boldsymbol{\Delta \mu}\downarrow$ & $\boldsymbol{\Delta \sigma}\downarrow$ & $\boldsymbol{\Delta P}\downarrow$ \\

\midrule
Cyclic order & 0.004 & 0.000 & 1.273 \\
Reverse order & 0.000 & 0.000 & 1.342 \\
2px noise & 1.304 & 0.002 & 0.176 \\
4px noise & 2.328 & 0.003 & 0.588 \\
\bottomrule
\end{tabular}
\end{table}

\paragraph{Network Details.}
To clarify the architectural design of our framework, Table~\ref{tab:net_arch} summarizes the detailed layer configurations of \emph{GuidanceNet} and \emph{DenoisingNet}. \emph{GuidanceNet} focuses on extracting multi-modal structural cues through a token-encoding paradigm and a Dual-Path Encoder, and further produces Gaussian Room Embeddings via an autoregressive prediction head. This design enables the model to jointly encode semantic, geometric, and boundary-level information for controllable generation and editing. In contrast, \emph{DenoisingNet} employs a time-aware transformer architecture that denoises polygonal representations conditioned on boundary and anchor constraints. Together, these two modules form the core of \emph{GRE-Diff}, with \emph{GuidanceNet} providing high-level room embeddings and \emph{DenoisingNet} ensuring accurate, constraint-aware vector generation.

\begin{figure}[t]
    \centering
    \includegraphics[width=\linewidth]{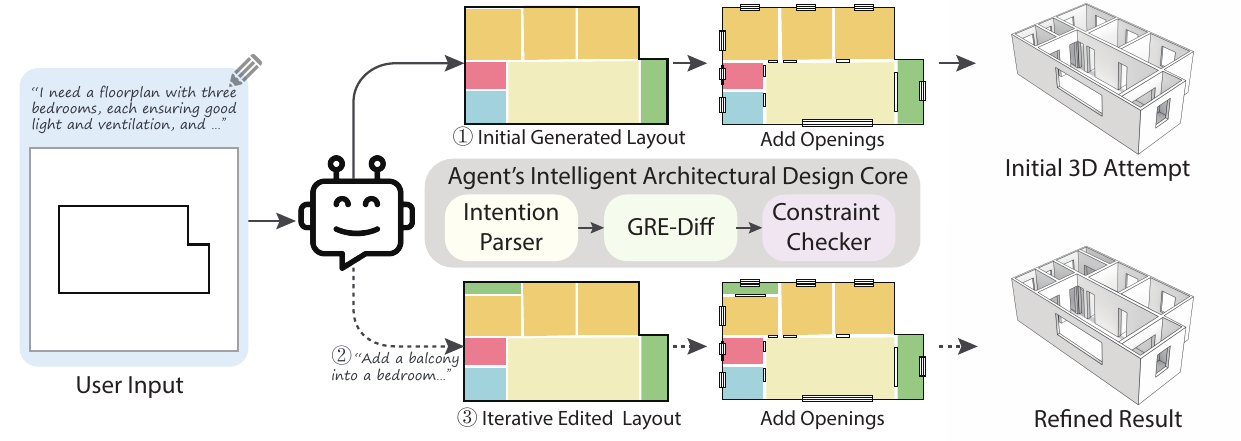} 
    \caption{The iterative workflow of the proposed AI architect agent. The agent processes multi-modal user inputs (natural language prompts and boundary constraints) through its {Intelligent Architectural Design Core}. This core acts as an autonomous central controller that coordinates three vital modules: an {Intention Parser} for extracting structured requirements, a generative model ({GRE-Diff}) for layout creation and modification, and a {Constraint Checker} for evaluating physical and architectural validity. Solid arrows delineate the initial generation cycle leading to the first 3D attempt. Dashed arrows highlight the agent's capacity for multi-turn iterative refinement, actively incorporating subsequent feedback (e.g., ``Add a balcony...'') to generate the refined final result.}
    \label{fig:agent_workflow}
\end{figure}

\paragraph{Limitation Cases.}
While our method demonstrates strong and stable performance in most scenarios, it still exhibits several limitations under challenging or atypical conditions. Even in these cases, however, the generated layouts often preserve reasonable geometric or semantic structure, highlighting the inherent robustness of \emph{GRE-Diff}.

\textit{(1) Sensitivity to Corner Point Density.} 
The model is sensitive to the density of corner points used to define the building boundary. Insufficient corner points may cause the interior space to be insufficiently covered, whereas an excessive number may lead to irregular or unstable room geometries. Even so, the generated layouts typically maintain overall topological continuity.

\textit{(2) Semantic Inconsistencies.}
Semantic assignments can sometimes become unstable, resulting in mismatches between a room’s geometry and its predicted functional type. %Although the semantic label may be incorrect, the geometric arrangement of rooms still tends to form a plausible and coherent spatial configuration.

\textit{(3) Dataset Bias and Wall Curvature.}
Due to the dataset’s bias toward rectilinear and low-curvature building boundaries, the model may exhibit reduced performance when presented with highly curved or free-form contours. Nevertheless, it still aligns room boundaries reasonably well with inclined or sloped edges, demonstrating partial generalization beyond the training distribution.

\textit{(4) Door and Window Placement.} 
Recent state-of-the-art vector-based floor plan generation methods increasingly focus on predicting room layouts rather than explicitly modeling doors or windows.
For example,~\cite{qin2024chathousediffusion,hu2025gsdiff} generate room geometries while leaving the placement of openings unspecified.
Earlier rule-based systems~\cite{wu2019Data-Driven} instead inserted doors along boundaries shared by adjacent rooms and placed windows on exterior walls using simple geometric heuristics.
Following this paradigm, our method also does not regress door or window geometry directly.
Instead, we prioritize producing accurate and well-structured room layouts, while openings are added afterward through a lightweight rule-based procedure.
This design emphasizes reliable room positioning and topology during generation, which is particularly beneficial for interactive editing where room placement is the primary operation.
Once the room configuration is stabilized, doors and windows can be deterministically derived from geometric relations (e.g., shared walls and exterior boundaries), leading to more consistent and controllable results.
This strategy avoids the error amplification that would occur if openings were predicted jointly with room geometry and removes the need for additional supervision or training objectives.

\paragraph{Future Extensions.}
Since \emph{GRE-Diff} already provides structured interfaces for both layout generation and editing, it can potentially be extended to an Agent-driven design framework, as shown in Fig.~\ref{fig:agent_workflow}.
In such a setting, a large language model (LLM) could act as a high-level planner that interprets user requirements, evaluates generated layouts, and iteratively invokes the editing interfaces to refine the design.
By leveraging the controllable operators provided by \emph{GRE-Diff}, the Agent could automatically formulate editing actions (e.g., adding rooms, adjusting spatial relations, or anchoring key regions) and explore alternative layouts.
This extension would enable autonomous layout refinement while maintaining precise geometric control, and may help resolve conflicting user constraints during the design process.

%\paragraph{Toward real-world architectural pipelines.}
%Our framework is also compatible with practical architectural design workflows.
%In real-world scenarios, structural elements such as load-bearing walls are typically fixed and must be preserved during layout design.
%The room-centric representation adopted in \emph{GRE-Diff} naturally supports this requirement by prioritizing the generation of stable room configurations.
%In principle, once the room layout is finalized, downstream modules could derive doors, windows, and elevation information using lightweight rule-based procedures.
%Such a pipeline would allow the global layout to be optimized first, while openings and other architectural details are inferred afterward from geometric relationships (e.g., shared walls or exterior boundaries).
%This separation simplifies local editing and avoids the complexity introduced by jointly optimizing openings and room geometry.

% \input{tab/tab_DenoisingNet}

\begin{figure*}[ht]
    \centering
    \includegraphics[width=0.85\linewidth]{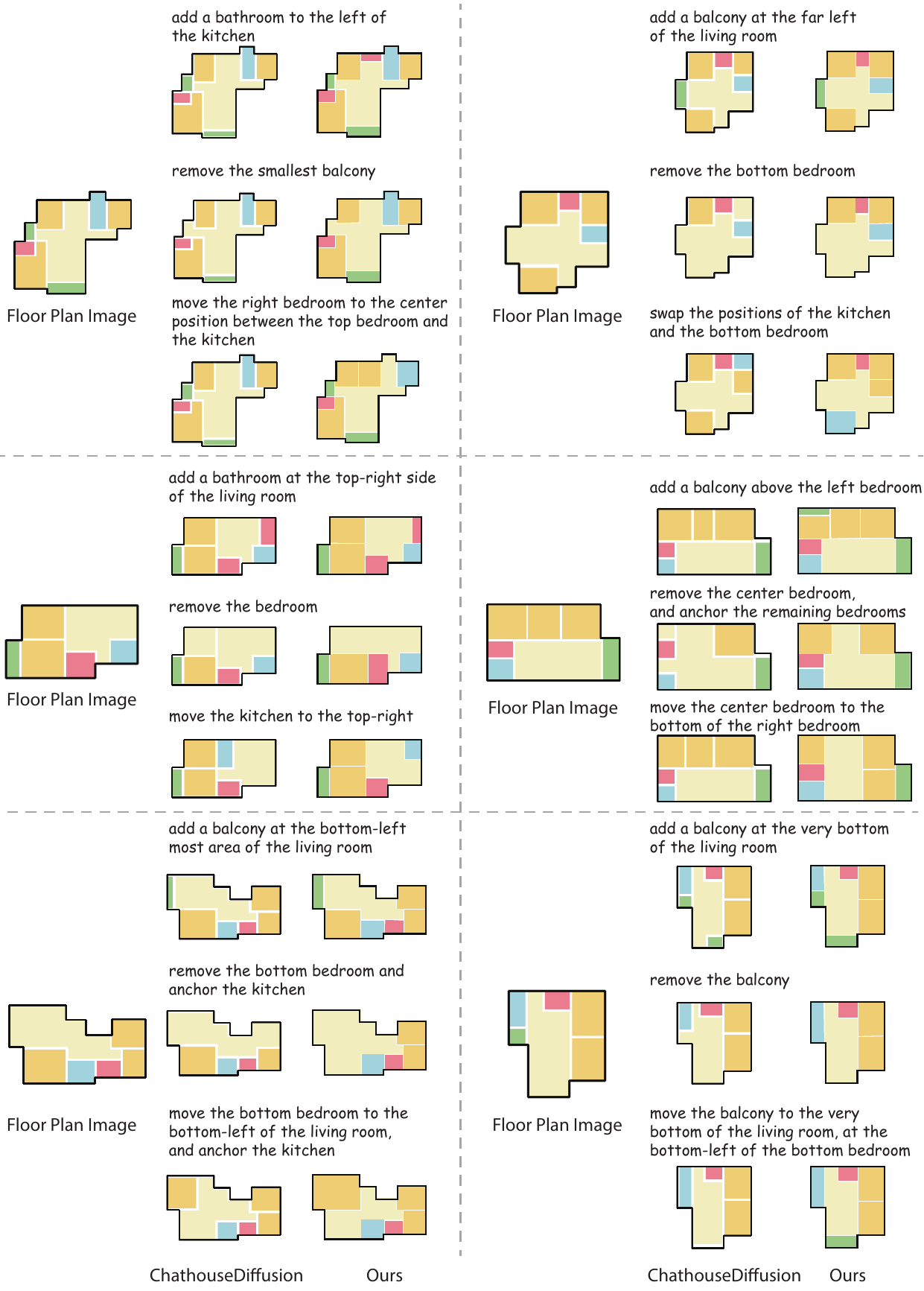}
    \caption{Comparison of floor plan editing effectiveness between ChatHouseDiffusion and our method. The images illustrate the results of various editing operations, including adding, removing, and moving rooms with specific LLM-based constraints. The left column shows the edits applied using ChatHouseDiffusion, while the right column shows the corresponding results using our method. The examples demonstrate how our method better maintains structural integrity and coherence during edits, especially when rooms are added or removed, or when fixed elements like the kitchen are moved. }%The results highlight the superior controllability and flexibility of our approach in generating floor plans under user-specified constraints.}
    \label{fig:test_editing}
    \end{figure*}

\begin{table}[t]
    \centering
    \renewcommand{\arraystretch}{1.2} 
    \setlength{\tabcolsep}{4pt}
    \captionsetup{font={small, stretch=1.1}}
    \caption{The architectures of \emph{GRE-Diff}.}\label{tab:net_arch}
    \scalebox{0.6}{
        \begin{tabular}{cccc}
            % \toprule
            \hline 
            \textbf{Architecture} & \textbf{Layer} & \textbf{SubLayer} & \textbf{Specification} \\
            % \midrule
            \hline 
            \hline 
            \multirow{9}{*}{\makecell{Condition \\ Token Encoder}} & \multirow{3}{*}{\makecell{Semantic \\ Token Encoder}} & semantic\_mlp0.weight & [512,256] \\
            & & semantic\_mlp1.weight & [256,256] \\
            & & semantic\_embed.weight & [256,256] \\
            \cline{2-4} 
            & \multirow{3}{*}{\makecell{Boundary \\ Token Encoder}} & boundary\_mlp0.weight & [512,256] \\
            & & boundary\_mlp1.weight & [256,256] \\
            & & boundary\_embed.weight & [256,256] \\
            \cline{2-4} 
            & \multirow{3}{*}{\makecell{Polygon \\ Token Encoder}} & polygon\_mlp0.weight & [512,256] \\
            & & polygon\_mlp1.weight & [256,256] \\
            & & polygon\_embed.weight & [256,256] \\
            \cline{1-4} 

            \multirow{9}{*}{\makecell{Dual-Path\\ Encoder}} & \multirow{7}{*}{\makecell{Polygon \\ Transformer\_i \\ (i=0,1)}} 
            & attn\_local\_boundary.weight & [256,768] \\
            & & attn\_global\_semantic.weight & [256,768] \\
            & & attn\_global\_polygon.weight & [256,768] \\
            & & attn\_local\_semantic.weight & [256,768] \\
            & & attn\_local\_polygon.weight & [256,768] \\
            & & linear1.weight & [256,1024] \\
            & & linear2.weight & [1024,256] \\
            \cline{2-4}
            & \multirow{2}{*}{Adapter} 
                & gen\_adapter\_linear & [256,256] \\
            &   & edit\_adapter\_linear & [256,256] \\
            % &   & gen\_adapter\_linear & [256,256] \\
            % &   & edit\_adapter\_linear & [256,256] \\
            \cline{1-4}

            \multirow{3}{*}{Auxiliary Head} & \multirow{3}{*}{-} & semantic\_predictor.weight & [256,10] \\
            & & mu\_predictor.weight & [256,2] \\
            & & sigma\_predictor.weight & [256,1] \\
            \cline{1-4}

            \multirow{10}{*}{\makecell{Auto-Regressive \\ Transformer}} & \multirow{4}{*}{\makecell{Decoder\_layer\_i \\ (i=0,1,...,11)}} 
            & to\_q.weight & [512,256] \\
            & & to\_k.weight & [512,256] \\
            & & to\_v.weight & [512,256] \\
            & & to\_out.weight & [256,512] \\
            \cline{2-4}
            & \multirow{6}{*}{Output Heads} 
            & to\_mu\_x\_linear0.weight & [256,256] \\
            & & to\_mu\_x\_linear2.weight & [256,50] \\
            & & to\_mu\_y\_linear0.weight & [256,256] \\
            & & to\_mu\_y\_linear2.weight & [256,50] \\
            & & to\_sigma\_linear0.weight & [256,256] \\
            & & to\_sigma\_linear2.weight & [256,50] \\
            \hline 
            \hline 
            \multirow{8}{*}{\makecell{Boundary Encoder}} 
            & \multirow{1}{*}{\makecell{ResNet50 Backbone}} 
                & - & - \\
            \cline{2-4}
            & \multirow{7}{*}{\makecell{MSDeformAttn\_i \\ (i=0,1,...,5)}} 
                & sampling\_offsets.weight & [256,256] \\
            &   & attention\_weights.weight & [256,128] \\
            &   & value\_proj.weight & [256,256] \\
            &   & output\_proj.weight & [256,256] \\
            &   & linear1.weight & [256,1024] \\
            &   & linear2.weight & [1024,256] \\
            \cline{1-4}
            \multirow{1}{*}{\makecell{Polygon Encoder}} & \multirow{1}{*}{-} & polygon\_embed.weight & [256,256] \\

            \cline{1-4} 
            \multirow{2}{*}{\makecell{Time Encoder}} & \multirow{2}{*}{-} & time\_embed0.weight & [128,512] \\
            & & time\_embed1.weight & [512,512] \\
            \cline{1-4} 

            \multirow{10}{*}{\makecell{Room Transformer \\ Decoder}} 
            
            & \multirow{10}{*}{\makecell{DeformTrans \\ Decoder\_i \\ (i=0,1,...,5)}} 
                & cross\_attn.sampling\_offsets.weight & [256,256] \\
            &   & cross\_attn.attention\_weights.weight & [256,128] \\
            &   & cross\_attn.value\_proj.weight & [256,256] \\
            &   & cross\_attn.output\_proj.weight & [256,256] \\
            &   & self\_attn.in\_proj\_weight & [256,768] \\
            &   & self\_attn.out\_proj.weight & [256,256] \\
            &   & self\_attn\_poly.in\_proj\_weight & [256,768] \\
            &   & self\_attn\_poly.out\_proj.weight & [256,256] \\
            &   & linear1.weight & [256,1024] \\
            &   & linear2.weight & [1024,256] \\
            \cline{1-4}

            \multirow{3}{*}{Output MLP} & \multirow{3}{*}{-} 
            & linear\_1.weight & [512,256] \\
            & & linear\_2.weight & [256,256] \\
            & & linear\_3.weight & [256,2] \\
            \hline 
            \vspace{-30pt}
        \end{tabular}}
\end{table}

\end{document}